\PassOptionsToPackage{table}{xcolor}
\documentclass[10pt, a4paper, copyright]{krafton-ai}

\usepackage[authoryear, sort&compress, round]{natbib}
\usepackage{tabularx}
\usepackage{wrapfig}
\usepackage[table]{xcolor}
\usepackage{booktabs}
\usepackage{tabularx}
\usepackage{ragged2e}
\usepackage{makecell}
\usepackage{xspace}
\usepackage{pifont}
\usepackage{fontawesome5}
\usepackage{subcaption}
\usepackage{graphicx}
\usepackage{multirow}
\usepackage[most]{tcolorbox}
\tcbuselibrary{listings,breakable}

\lstdefinestyle{bt}{
  basicstyle=\ttfamily\small,
  backgroundcolor=\color{black!6},
  frame=single,
  rulecolor=\color{black!30},
  frameround=tttt,
  framesep=6pt,
  numbers=left,
  numberstyle=\tiny\color{black!50},
  numbersep=10pt,
  xleftmargin=2.2em,
  showstringspaces=false,
  columns=fullflexible,
  keepspaces=true,
  keywordstyle=\bfseries\color{teal!70!black},
  morekeywords={selector,sequence,condition,task},
}

\def\tabref#1{Table~\ref{#1}}

\NewTColorBox[auto counter]{example}{ O{!htbp} m O{} }{
  enhanced,
  breakable,
  float*,
  floatplacement={#1},
  width=\textwidth,
  colback=gray!5!white,
  colframe=gray!75!black,
  title={Example~\thetcbcounter: #2},
  enlarge top by=5mm,
  enlarge bottom by=5mm,
  label={#3}
}

\newif\ifreview
\reviewtrue   

\ifreview
  
\else
  
\fi

\newcommand{\RaonSpeech}{Raon-Speech\xspace}
\newcommand{\RaonChat}{Raon-SpeechChat\xspace}
\newcommand{\SIL}{\texttt{SIL}\xspace}
\newcommand{\EPAD}{\texttt{BOW}\xspace}
\newcommand{\PAD}{\texttt{PAD}\xspace}
\newcommand{\BC}{\texttt{BC}\xspace}

\uselogo{}

\title{\centering \RaonSpeech Technical Report}

\usepackage[useregional=false]{datetime2}
\DTMsetstyle{iso}
\paperdate{\DTMtoday}

\author{\centering \quad \normalsize{KRAFTON$^{\dagger}$}}

\makeatletter
\renewcommand{\maketitle}{
  \bgroup\setlength{\parindent}{0pt}
  \begin{adjustwidth}{0pt}{24pt}
    \begin{flushleft}
      {\raggedright \titlefont \@title\par}
      \vskip11pt
      {\centering\Large\textbf{\@author}\par}
      \vskip4pt
      \vskip12pt
    \end{flushleft}
  \end{adjustwidth}
  \egroup
  {\abscontent}
  \begin{center}
    \includegraphics[width=\linewidth]{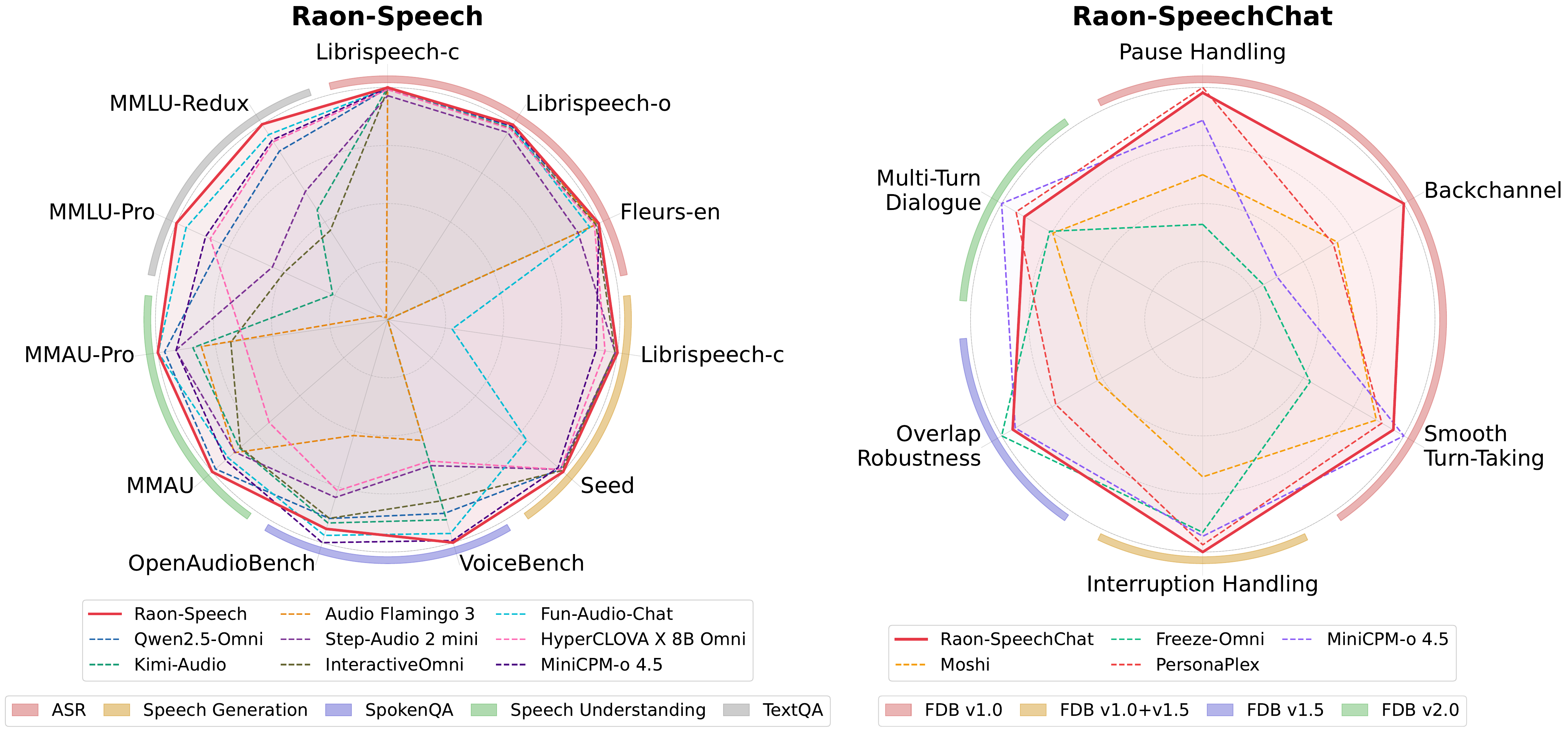}
    \captionof{figure}{Overall performance comparison of Raon-Speech and Raon-SpeechChat against baseline models across diverse benchmarks. \textbf{(Left)} Raon-Speech evaluated on English speech and text benchmarks spanning automatic speech recognition (ASR), speech generation, spoken question answering (SpokenQA), speech understanding, and text question answering (TextQA). See Appendix~\ref{app:kor_overall_performance} for performance on Korean benchmarks. \textbf{(Right)} Raon-SpeechChat evaluated on the Full-Duplex-Bench (FDB) covering pause handling, backchannel, smooth turn-taking, interruption handling, overlap robustness, and multi-turn dialogue. All scores are zero to max normalized per benchmark axis.}
  \end{center}
  \thispagestyle{firststyle}
}
\renewcommand{\abscontent}{
  \begin{tcolorbox}[
    enhanced,
    frame hidden,
    colback=KraftonLightGray,
    arc=4pt,
    left=12pt, right=12pt, top=12pt, bottom=12pt,
    before skip=0pt, after skip=0pt
  ]
  {\absfont \theabstract}
  \@ifundefined{@keywords}{}{
    \vskip1em \noindent \keywordsfont Keywords: \@keywords}
  \vskip0.8em
  \noindent{\textcolor{black}{\faGithub~GitHub:}\enspace\textcolor{cyan!60!black}{\href{https://github.com/krafton-ai/Raon-Speech}{https://github.com/krafton-ai/Raon-Speech}}}\\[0.3em]
  \noindent\makebox[0pt][l]{\hspace{-1.5pt}\raisebox{-3pt}{\includegraphics[height=1.3em]{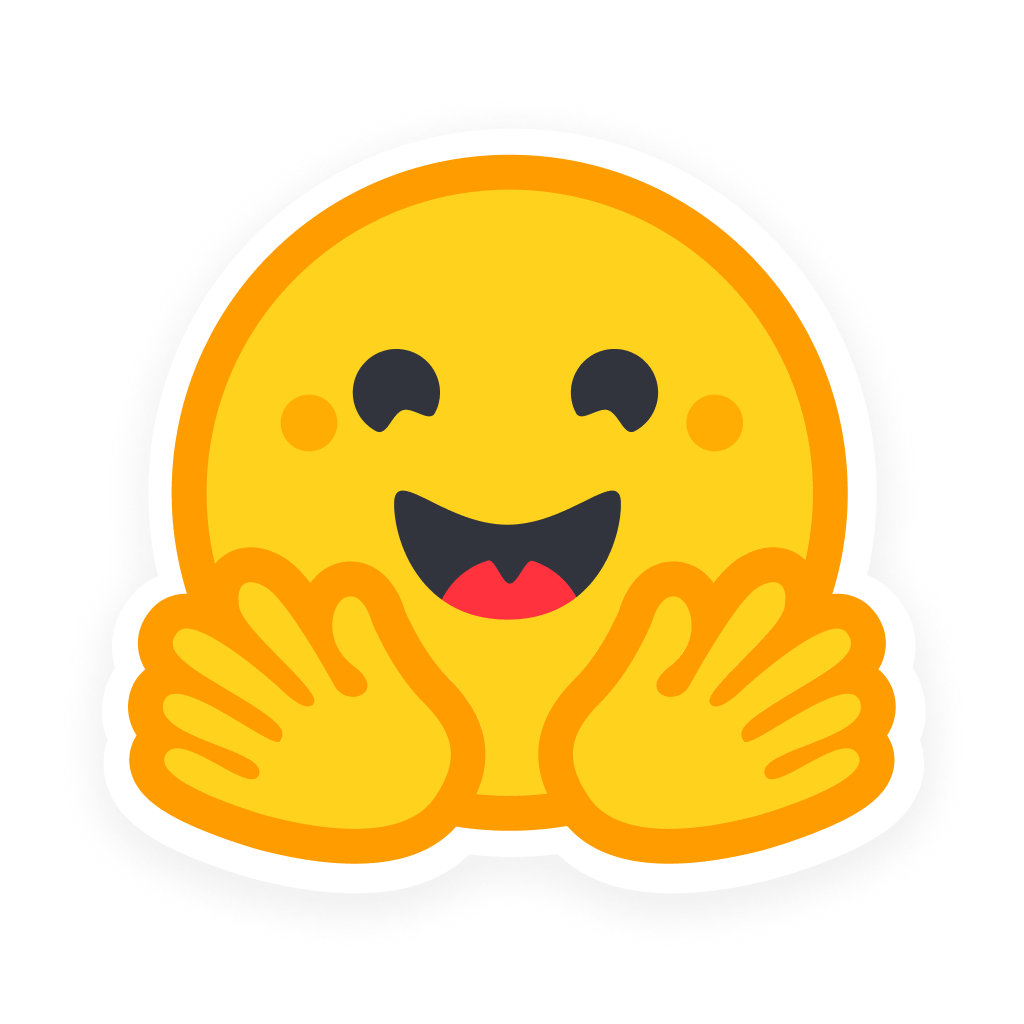}}\hspace{2pt}HuggingFace:\enspace\textcolor{cyan!60!black}{\href{https://huggingface.co/collections/KRAFTON/raon}{https://huggingface.co/collections/KRAFTON/raon}}}
  \end{tcolorbox}
}
\makeatother

\begin{abstract}
We present \textbf{\RaonSpeech}, a top-performing 9B-parameter speech language model~(SpeechLM) for English and Korean speech understanding, answering, and generation, and \textbf{\RaonChat}, a high-performing full-duplex extension for natural real-time conversation.
\RaonSpeech successfully transforms a pre-trained LLM into a SpeechLM that both understands and generates speech while preserving strong text capabilities.
It trains on 1.38M hours of highly curated English and Korean speech and text datasets with the following training stages: (1)~speech modules alignment, (2)~end-to-end SpeechLM pre-training with knowledge distillation, and (3)~multi-task preference optimization-based post-training.
Across 42 English and Korean speech and text benchmarks, \RaonSpeech establishes the strongest overall profile on speech-centric tasks in our comparison against eight similarly sized recent audio foundation models, including Qwen2.5-Omni and Fun-Audio-Chat, while preserving strong text question answering performance.
Building upon it, \RaonChat enables natural full-duplex conversation by continual training on {119}K hours of time-aligned real and synthetic dialogue data.
It proceeds through three complementary training stages:
(1)~causal encoder adaptation, (2)~full-duplex pre-training, (3)~full-duplex fine-tuning for voice and role-control.
On multiple full-duplex benchmarks, \RaonChat shows its clearest strengths on the turn-taking and interruption-sensitive behaviors covered by FDB v1.0, and remains competitive across the broader full-duplex evaluation suite.
We open-source all model checkpoints, the training and inference pipeline, and an interactive demo.
\end{abstract}

\begin{document}

\maketitle
{\let\thefootnote\relax\footnotetext[1]{$^\dagger$ The complete list of authors is in the Authorship and Credit Assignment section.}}

\section{Introduction}\label{sec:intro}
Speech plays a central role in human cognition~\citep{hickok2012computational}. Through spoken language, humans can perceive their surroundings, express intentions, and interact with one another in real time, giving rise to complex, dynamic, and richly social systems. The centrality of speech is increasingly reflected in modern computing, from in-vehicle voice assistants and game-playing agents to voice-based robot control and computer use~\citep{clark2019state, arora2025landscape}. Accordingly, there is a growing interest in developing interactive speech-language systems that support more human-like communication. Unlike text, speech carries not only linguistic content but also prosody, timing, and turn-taking cues that are essential for natural interaction.

Speech language models~(SpeechLMs)
are emerging as the most promising path toward this goal.
By extending the strong language capabilities of large language models (LLMs) to the speech modality, they enable natural and high-quality spoken interaction~\citep{defossez2024moshi,goel2025audioflamingo,team2025fun}. However, a gap remains between current models and practical deployment. Lightweight models (\textit{i.e.}, under 10B parameters) still struggle to deliver high-quality multilingual speech interaction beyond English, while full-duplex models remain limited
in temporal awareness and interaction naturalness, particularly in settings that require
delicate real-time communication such as dynamic games~\citep{chang2025game}. Practical deployment further requires low latency, robust interruption handling, and coherent turn-taking, all of which remain challenging for current SpeechLMs.

In this paper, we present \textbf{\RaonSpeech}, a 9B-parameter SpeechLM for English and Korean speech understanding, answering, and generation, and \textbf{\RaonChat}, its extension for natural real-time conversation via full-duplex interaction. \RaonSpeech augments a pre-trained LLM backbone with speech understanding and generation modules, acquiring new speech modality capabilities through a staged training recipe while preserving the backbone's original text proficiency. \RaonChat further incorporates three complementary changes for real-time simultaneous listening and speaking: (1) a causal encoder for streaming input; (2) a token-level interleaved sequence over user speech, assistant text, and assistant speech with word-level alignment; and (3) state modeling that separates when to speak from what to say, enabling controllable interaction timing and behavior.

Through extensive experiments on 42 English and Korean speech and text benchmarks, \RaonSpeech establishes the strongest speech-centric profile in our comparison against eight similarly sized recent audio foundation models. In English, its clearest gains are in spoken question answering, speech understanding, and generated-speech intelligibility, as reflected by the highest VoiceBench average, the best MMAU and MMAU-Pro scores, and the lowest WER on LibriSpeech~\citep{panayotov2015librispeech} and Seed-TTS-Eval~\citep{anastassiou2024seedtts}; it also preserves strong text capability, achieving the best MMLU-Pro~\citep{wang2024mmlupro} and MMLU-Redux results. In Korean, the gains are broader and stronger: \RaonSpeech achieves the best CER on all ASR and speech-generation benchmarks, the best KVoiceBench, KOpenAudioBench, and KMMAU scores, and the best KMMLU-Pro and KMMLU-Redux results. For readability, the main result tables report aggregate VoiceBench/OpenAudioBench and KVoiceBench/KOpenAudioBench scores, while Appendix~\ref{app:spokenqa_details} provides the per-benchmark spoken question answering breakdowns.
\RaonChat further shows the strongest overall ability on the turn-taking and interruption-sensitive behaviors covered by FDB v1.0, while remaining competitive under overlapped speech in the broader full-duplex evaluation suite.

Our contributions are summarized as follows:
\begin{itemize}
  \item We introduce \textbf{\RaonSpeech}, a 9B-parameter SpeechLM, and show the strongest speech-centric profile in our comparison against eight similarly sized recent audio foundation models across 42 English and Korean speech and text benchmarks.
  \item We propose \textbf{\RaonChat}, a full-duplex model that enables natural real-time conversation in various challenging scenarios like games.
  \item We release 3 Korean speech benchmarks, KVoiceBench, KOpenAudioBench, and KMMAU, which are tailored to Korean speech and culture.
  \item We open-source all model checkpoints, the inference pipeline, and an interactive demo.
\end{itemize}

\section{Model Architecture}\label{sec:model}
\subsection{\RaonSpeech}
\begin{figure*}[t]
    \centering
    \includegraphics[width=0.75\textwidth,page=1]{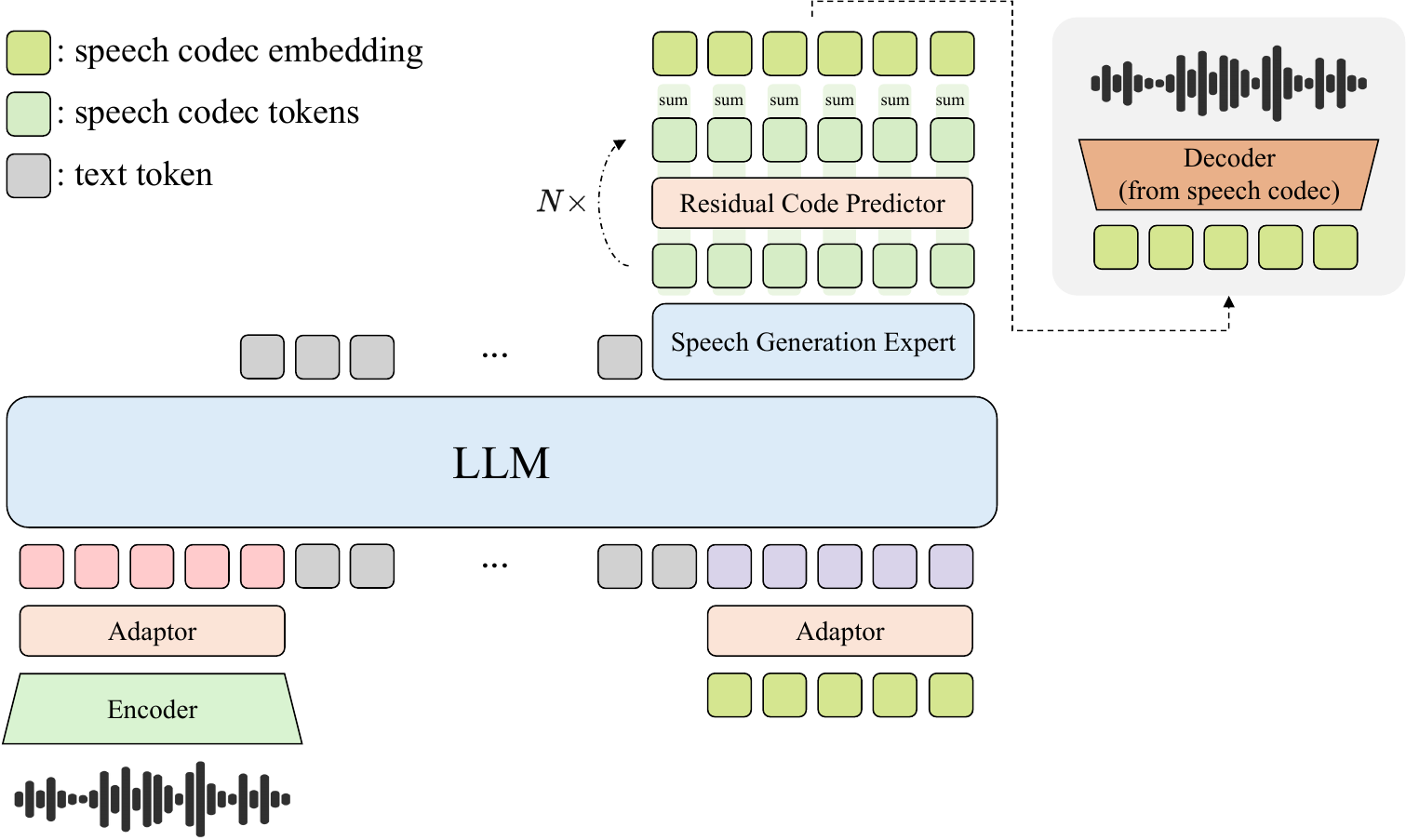}
    \caption{\textbf{Overview of \RaonSpeech and \RaonChat.} \RaonSpeech extends the backbone LLM to support speech understanding and generation. On the speech understanding side, input speech is passed through the speech encoder and adaptor to obtain speech embeddings, which are then fed into the LLM as input representations. On the speech generation side, the codec embedding from the previous step is first mapped through the output adaptor into the LLM input space. The speech generation expert then predicts the semantic token from the LLM hidden states, after which the residual code predictor (RCP) predicts the acoustic tokens residual-wise for 15 steps. Dequantizing the predicted code at each depth yields codec hidden states, and summing the semantic and all acoustic depths produces the codec embedding. This embedding is passed through the codec decoder to synthesize speech, and is also used as the input to the output adaptor for the next generation step.}
    \label{fig:raonspeech_overview}
\end{figure*}

Figure~\ref{fig:raonspeech_overview} illustrates the overall architecture of \RaonSpeech, which extends a pre-trained backbone LLM to support both speech understanding and speech generation. We adopt \texttt{Qwen3-VL-8B-Instruct}~\citep{bai2025qwen3vltechnicalreport} as the backbone LLM for its strong multilingual text capabilities. The \textit{speech understanding modules} consist of a speech encoder and an input adaptor, and the \textit{speech generation modules} consist of an output adaptor, a speech generation expert, a residual code predictor (RCP), and a speaker encoder. The module-wise parameter breakdown of \RaonSpeech is provided in Appendix~\ref{app:param_breakdown} (Table~\ref{tab:param_counts}), while its detailed architectural configuration is given in Appendix~\ref{app:model_config} (Table~\ref{tab:arch_compact}).

\paragraph{Speech understanding modules.}
A speech encoder, initialized from the pre-trained AuT model~\citep{qwen3_asr_technical_report} for its strong multilingual speech representations, first extracts features from input speech at a 12.5 Hz token rate. A randomly initialized input adaptor, implemented as a 2-layer MLP with GELU activation following~\citep{liu2023visual}, then projects the encoder outputs into the LLM embedding space. The adapted speech embeddings are inserted into the LLM input sequence, allowing the backbone to process speech through the same interface used for text. This design enables the backbone to reuse its pre-trained language capability while delegating speech-specific feature extraction to the encoder and adaptor. To stabilize alignment between speech and text representations, we apply RMSNorm~\citep{NEURIPS2019_1e8a1942} after the adaptor with its weight set to a small scale of 0.02, such that the norm of speech embeddings matches the norm of the LLM embeddings at initialization.

\paragraph{Speech generation modules.}
 We use the Mimi codec~\citep{defossez2024moshi}, an RVQ-based neural speech codec~\citep{zeghidour2021soundstream, defossez2022high} designed for streaming generation. Mimi uses 32 residual codebooks, of which we retain the first 16 to balance efficiency and generation quality in real-time settings. At each generation step, \RaonSpeech predicts 16 codec tokens, consisting of 1 semantic token at the first residual depth and 15 acoustic tokens at the subsequent depths. A randomly initialized output adaptor, sharing the same architecture as the input adaptor, maps the codec embedding from the previous step into the input space of the backbone LLM. From the resulting backbone hidden states, a randomly initialized four-layer decoder-only Transformer speech generation expert, whose hidden size is smaller than that of the backbone, predicts the semantic token. A 15-depth RCP, initialized from \texttt{Qwen3-Omni-30B-A3B-Instruct}~\citep{xu2025qwen3} to accelerate convergence, then predicts the remaining acoustic tokens across residual depths. For speaker identity control, we condition the model on speaker embeddings extracted using a speaker encoder initialized from \texttt{speechbrain/spkrec-ecapa-voxceleb}~\citep{desplanques2020ecapa, dawalatabad2021ecapa}. Specifically, a random chunk from the target speech, with duration ranging from 2 seconds to 8 seconds for \RaonSpeech (10 seconds for \RaonChat), is encoded and inserted into the LLM input sequence to condition the speaker identity of the generated speech.

\subsection{\RaonChat}
\begin{figure*}[t]
    \centering
    \includegraphics[width=0.70\textwidth,page=1]{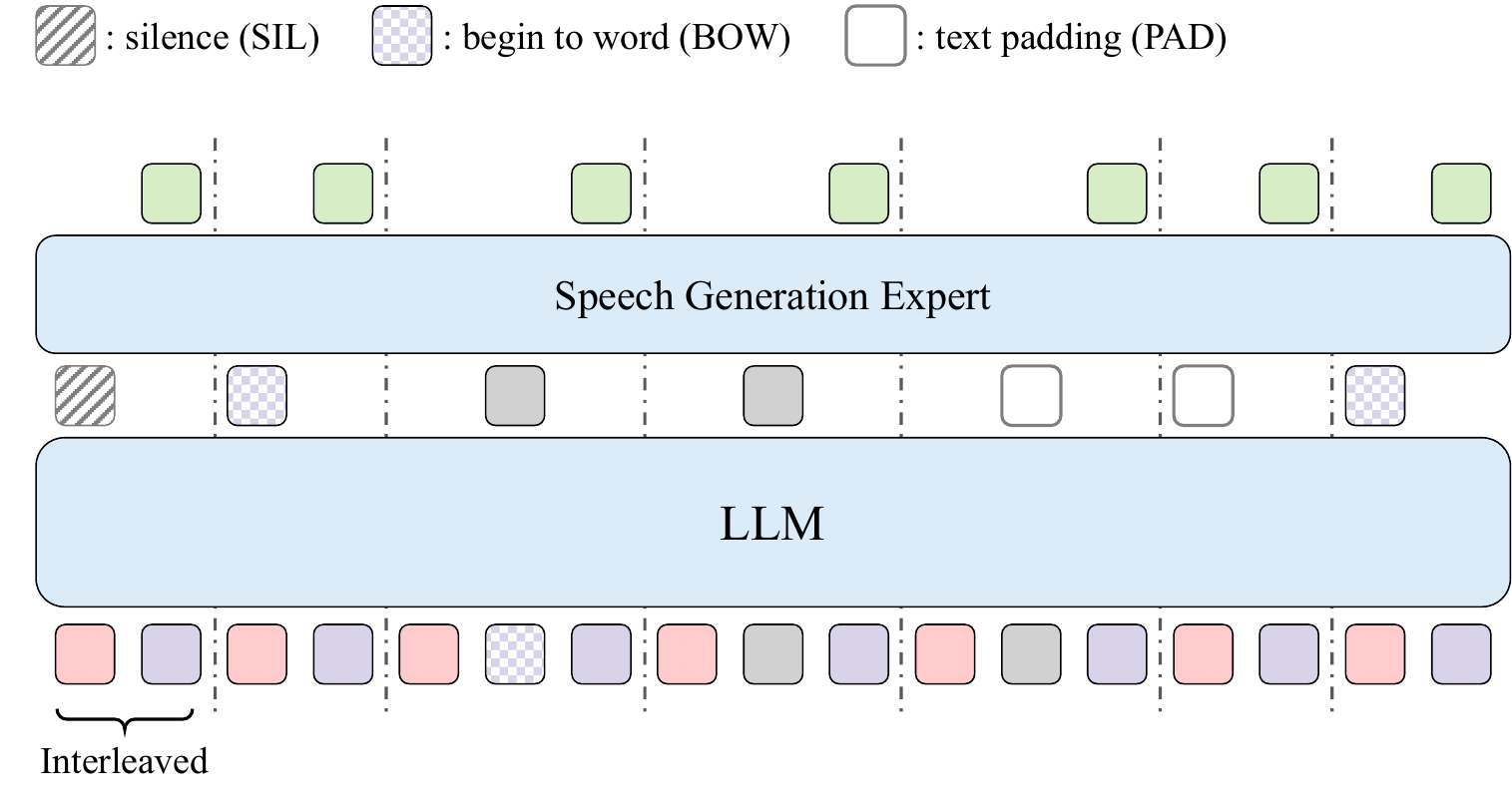}
    \caption{\textbf{\RaonChat overview.}
    \RaonChat uses a causal speech encoder to obtain streaming user-speech embeddings and is trained on an interleaved token sequence of user speech, assistant text, and assistant speech. During duplex generation, it continuously listens while speaking in parallel. Dashed vertical lines indicate aligned time boundaries. In the example, the assistant remains silent while listening (\SIL), then begins speaking with \EPAD. The text token is generated first; since speech generation lasts longer, \PAD\ tokens are emitted until the spoken word is completed, after which another \EPAD\ marks the next word onset.}
    \label{fig:raonchat_overview}
\end{figure*}

We describe architectural and sequence-level modifications that extend \RaonSpeech to real-time full-duplex conversation.
Figure~\ref{fig:raonchat_overview} illustrates the token-sequence interleaving used in \RaonChat.
Specifically, supporting simultaneous listening and speaking requires the model to process
streaming user audio while generating assistant speech in parallel. To this end, we introduce three modifications: (1) a causal speech encoder for streaming input;
(2) token-sequence interleaving over user speech, assistant text, and assistant speech;
(3) explicit interaction-state modeling that separates when-to-speak from what-to-say, enabling control over interaction timing and behavior.
The module-wise parameter breakdown of \RaonChat is provided in Appendix~\ref{app:param_breakdown} (Table~\ref{tab:param_counts}), while its detailed architectural configuration is given in Appendix~\ref{app:model_config} (Table~\ref{tab:arch_compact}).

\paragraph{Causal speech encoder.} Starting from \RaonSpeech, we replace the non-causal AuT encoder with the causal speech encoder of \texttt{
Voxtral-Mini-4B-Realtime-2602}~\citep{voxtral_realtime}.
The encoder uses causal attention and is designed for native streaming operation, allowing user audio to be processed without future context.
Combined with sliding window attention, it supports efficient long-form streaming by restricting attention to a fixed window, allowing continuous transcription with controllable latency.
To facilitate replacing the non-causal encoder with the causal one, we introduce a re-adaptation training stage (See Section~\ref{sec:training}).

\paragraph{Full-duplex sequence design.}
To model simultaneous speaking and listening, \RaonChat is trained on a single autoregressive sequence that interleaves user speech, assistant text, and assistant speech, rather than modeling them in parallel as in Moshi~\citep{defossez2024moshi}.
The user speech stream is continuously encoded and consumed by the backbone, while the assistant side is modeled through interleaved text and speech tokens within the same autoregressive sequence.
Assistant text and assistant speech are aligned at the word level, and temporal consistency is maintained with padding (\PAD) when the number of text tokens is smaller than the number of speech tokens.
This design allows recognition, language planning, and speech generation to be handled within a unified token-level modeling framework.

\paragraph{State modeling.}
We model the assistant’s conversational behavior using two states, \textit{listening} and \textit{speaking}, and govern transitions between them using special tokens predicted at each frame. In contrast to Moshi, which represents both silence and other non-text positions using a single \PAD token, \RaonChat introduces a dedicated silence token, \SIL (silence), to explicitly encode silent listening behavior. In the \textit{listening} state, predicting \SIL keeps the assistant silent, while predicting a speech-onset token moves the model to the \textit{speaking} state.

To further separate interaction timing from content generation, we introduce \EPAD (beginning of word), a special token emitted immediately before each assistant text token. Rather than carrying lexical content itself, \EPAD\ indicates that the model is about to produce a new word, thereby marking the boundary between deciding to speak and specifying what to say. Subsequent assistant text tokens then encode the actual response content. This design disentangles interactional behavior from linguistic content, allowing the model to learn \textit{when to speak} and \textit{what to say} through different token types. Once in the \textit{speaking} state, the model continues generating assistant text and speech until predicting \SIL, which returns it to the \textit{listening} state.

Finally, we use a separate token, \BC, for backchannels instead of reusing \EPAD. This allows the model to distinguish short listener responses from ordinary turn initiations more clearly during training, and provides explicit control over backchannel usage at inference time, including disabling backchannels altogether or adjusting their frequency.

\paragraph{Text lookahead.} To improve the stability and accuracy of speech generation, we introduce text lookahead so that, once the assistant begins speaking, text is generated ahead of speech. This reduces semantic drift between the evolving language plan and the generated speech tokens, and provides a more stable textual target for subsequent speech generation. In practice, text lookahead is particularly important in full-duplex settings, where speech must be generated incrementally under tight latency constraints.

\section{Training}\label{sec:training}
\subsection{\RaonSpeech}


\newcommand{\audio}{\texttt{\textless audio\textgreater}}
\newcolumntype{Y}{>{\raggedright\arraybackslash}X}

\begin{table*}[h!]
\centering
\footnotesize
\setlength{\tabcolsep}{5pt}
\renewcommand{\arraystretch}{1.14}
\caption[Representative pretraining task formats used in \RaonSpeech.]{
Representative pretraining task formats used in \RaonSpeech, shown with simplified chat-template-rendered examples. The tasks include STT (transcription), TTS (speech synthesis), SpeechQA and SpokenQA (speech-based question answering), and TextQA (text-based question answering). Audio spans are represented as \audio\ for readability.}
\label{tab:pretraining-task-examples}
\begin{tabularx}{\textwidth}{@{}l >{\raggedright\arraybackslash}p{0.16\textwidth} Y@{}}
\toprule
\textbf{Task} & \textbf{Input $\rightarrow$ Output} & \textbf{Simplified rendered example} \\
\midrule

\multirow{3}{*}{STT} 
& \multirow{3}{*}{\shortstack[l]{speech $\rightarrow$ \\ text}} 
& \textbf{User:} \audio\ Transcribe the audio into text. \newline
  \textbf{Assistant:} First, let's speak about our possible futures and how those are shaped by many agents of change. \\
\midrule

\multirow{3}{*}{TTS} 
& \multirow{3}{*}{\shortstack[l]{text $\rightarrow$ \\ speech}} 
& \textbf{User:} Speak the following text: There are pulls of the future caused by agents of change, such as social, technological, environmental, economic, and political. \newline
  \textbf{Assistant:} \audio \\
\midrule

\multirow{3}{*}{SpeechQA} 
& speech context + \newline text question $\rightarrow$ \newline text answer
& \textbf{User:} \audio \newline
  \textbf{User:} How old is the speaker? \newline
  \textbf{Assistant:} 16 \\
\midrule

\multirow{3}{*}{SpokenQA} 
& spoken context + \newline spoken question $\rightarrow$ \newline text answer
& \textbf{User:} \audio \newline
  \textbf{Assistant:} Dry palms will produce the most heat when rubbed together because the lack of moisture or lubrication increases friction between the surfaces. \\
\midrule

\multirow{3}{*}{TextQA}  
& text context + \newline text question $\rightarrow$ \newline text answer
& \textbf{User:} What kind of company is Krafton? \newline
  \textbf{Assistant:} Krafton is a game development and publishing company based in South Korea. \\
\bottomrule
\end{tabularx}
\end{table*}

\begin{table*}[t]
\centering
\scriptsize
\setlength{\tabcolsep}{3.5pt}
\renewcommand{\arraystretch}{1.10}
\caption{Stage-wise training setup of \RaonSpeech and \RaonChat. The table summarizes the main data/tasks and key optimization settings for each stage. \RaonSpeech\ consists of understanding alignment, generation alignment, pre-training with KD, and post-training, while \RaonChat\ consists of causal alignment, causal full training, full-duplex training, and two stages of full-duplex fine-tuning.}
\label{tab:training-hparams}
\begin{tabularx}{\textwidth}{@{}
    >{\raggedright\arraybackslash}p{0.14\textwidth}
    >{\raggedright\arraybackslash}p{0.16\textwidth}
    >{\raggedright\arraybackslash}p{0.16\textwidth}
    >{\raggedright\arraybackslash}p{0.25\textwidth}
    >{\raggedright\arraybackslash}p{0.07\textwidth}
    >{\raggedright\arraybackslash}X
@{}}
\toprule
\textbf{Stage} & \textbf{Main data / tasks} & \textbf{Objective} & \textbf{Optimization} & \textbf{Training steps} & \textbf{Batch / length} \\
\midrule

\multicolumn{6}{@{}c@{}}{\textbf{\RaonSpeech}} \\
\midrule

\makecell[l]{Understanding\\alignment} &
\makecell[l]{STT, SpeechQA,\\SpokenQA} &
CE &
\makecell[l]{LR $1.5\times10^{-4}\rightarrow1.5\times10^{-5}$;\\cosine; 950-step warmup} &
13.5k &
\makecell[l]{batch 128;\\len 24{,}576} \\
\midrule

\makecell[l]{Generation\\alignment} &
TTS &
CE &
\makecell[l]{LR $1\times10^{-4}\rightarrow1\times10^{-5}$;\\cosine; 6{,}550-step warmup} &
65.5k &
\makecell[l]{batch 64;\\len 8{,}192} \\
\midrule

\makecell[l]{Pre-training\\with KD} &
All tasks &
\makecell[l]{CE + KL;\\(on-policy KD)} &
\makecell[l]{LR $1.2\times10^{-5}\rightarrow1\times10^{-7}$;\\cosine; 4{,}000-step warmup} &
60k &
\makecell[l]{batch 112;\\len 8{,}192} \\
\midrule

Post-training &
\makecell[l]{Curated SFT\\(all tasks)} &
\makecell[l]{CE + SimPO} &
\makecell[l]{LR $1\times10^{-6}$;\\constant; 112-step warmup} &
800 &
\makecell[l]{batch 96;\\len 8{,}192} \\
\midrule

\multicolumn{6}{@{}c@{}}{\textbf{\RaonChat}} \\
\midrule

\makecell[l]{Causal\\alignment} &
\makecell[l]{Same as\\\RaonSpeech} &
CE &
\makecell[l]{LR $1.5\times10^{-4}\rightarrow1.5\times10^{-5}$;\\cosine; 700-step warmup} &
10k &
\makecell[l]{batch 128;\\len 12{,}288} \\
\midrule

\makecell[l]{Causal full\\training} &
\makecell[l]{Same as\\\RaonSpeech} &
CE &
LR $2\times10^{-6}$ &
10k &
\makecell[l]{batch 160;\\len 8{,}192} \\
\midrule

\makecell[l]{Full-duplex\\training} &
\makecell[l]{Full-duplex + \\10\% of all tasks} &
CE &
\makecell[l]{LR $5\times10^{-6}\rightarrow5\times10^{-7}$;\\cosine; 1{,}250-step warmup} &
25k &
\makecell[l]{batch 128;\\len 4{,}096} \\
\midrule

\makecell[l]{Full-duplex\\fine-tuning I} &
\makecell[l]{High-quality\\conversation} &
CE &
\makecell[l]{LR $3\times10^{-6}\rightarrow3\times10^{-7}$;\\cosine; 250-step warmup} &
5k &
\makecell[l]{batch 64;\\len 4{,}096} \\
\midrule

\makecell[l]{Full-duplex\\fine-tuning II} &
\makecell[l]{High-quality\\synthetic data} &
\makecell[l]{CE; \EPAD\,$\rightarrow$\,\BC;\\$\times 50$ CE on \BC} &
Same as Stage I &
5k &
Same as Stage I \\
\bottomrule
\end{tabularx}
\end{table*}

The training of \RaonSpeech consists of three stages: alignment of speech modules, end-to-end SpeechLM training with knowledge distillation, and preference-based post-training. Across these stages, our goals are twofold: (1)~to equip the backbone LLM with speech understanding and generation capabilities, and (2)~to preserve its pre-existing text capabilities during speech adaptation. The task coverage spans speech-to-text (STT), text-to-speech~(TTS), speech question answering~(SpeechQA), spoken question answering~(SpokenQA), and text question answering~(TextQA); definitions and representative input-output examples are provided in Table~\ref{tab:pretraining-task-examples}. The training hyperparameters for each stage are summarized in Table~\ref{tab:training-hparams}.

 \paragraph{Alignment stage.} In this stage, the backbone LLM is kept frozen and only the newly introduced speech understanding and generation modules are trained, following standard practice in multimodal LLM alignment~\citep{liu2023visual, freezeomni2024, goel2025audioflamingo}. The goal is to align their representations with the backbone's embedding space before updating the backbone itself. Throughout this and all subsequent \RaonSpeech training stages, input audio is segmented into 8-second chunks and converted into token sequences.

\textit{Speech understanding alignment.} We train only the input adaptor while keeping the speech encoder and all other modules frozen. The training data consists of STT, SpeechQA, and SpokenQA
  samples in English and Korean. To improve robustness to real-world acoustic conditions, we apply  on-the-fly audio augmentation to user speech inputs following Moshi~\citep{defossez2024moshi}. Specifically, we add noise sampled from multiple corpora~\citep[DNS-Challenge]{dubey2024icassp} at SNRs ranging from $-30$ to $6$\,dB, apply synthetic reverberation, simulate channel distortion through response filtering, and degrade bandwidth. We trained for 13,500 iterations with a global batch size of 128 and a packed sequence length of 24,576 tokens. We used the AdamW optimizer~\citep{loshchilov2017decoupled} with a cosine learning rate schedule with a peak of 1.5e-4, a minimum of 1.5e-5, and a warmup of 950 steps.

\textit{Speech generation alignment.} Both the backbone LLM and the speech codec are kept frozen, while the speech generation modules, namely the output adaptor, the speech generation expert, and the RCP, are trained on English and Korean TTS data. Speaker embedding is enabled with a dropout rate of 0.2. We train for 65,500 iterations with a global batch size of 64 and a packed sequence length of 8,192 tokens. We use the AdamW optimizer with a cosine learning rate schedule, a peak learning rate of 1e-4, a minimum of 1e-5, and a warmup of 6,550 steps.

\paragraph{End-to-end pre-training with knowledge distillation.} The goal of this stage is to jointly train the backbone LLM with the speech modules, enabling more accurate speech understanding and generation while preserving its original text capabilities. We train on all five tasks: STT, TTS, SpeechQA, SpokenQA, and TextQA. Among these, TextQA is included specifically to mitigate forgetting of the backbone LLM's original text capability, whose ground-truth responses are generated by the backbone LLM ensuring consistency with its original output distribution. All modules are trainable except the speech encoder, speech codec, and speaker encoder; we find that freezing these encoder-side components leads to better performance and more stable optimization than fine-tuning them jointly. We train for 60,000 iterations with a global batch size of 112 and a packed sequence length of 8,192 tokens. We use the AdamW optimizer with a cosine learning rate schedule, a peak learning rate of 1.2e-5, a minimum of 1e-7, and a warmup of 4,000 steps.

The training objective is a weighted sum of cross-entropy loss and knowledge distillation losses with equal weights. SFT is applied to all five tasks, while knowledge distillation is applied to SpeechQA, SpokenQA, and TextQA. For knowledge distillation, we use two modality-dependent teachers. For audio inputs, we adopt a self-distillation approach: the teacher is the model itself conditioned on the corresponding text transcript, which explicitly encourages alignment between speech and text representations~\citep{wang2025cross, hu2026cord}. For text inputs, the teacher is the backbone LLM before pre-training, which helps mitigate catastrophic forgetting. Together, the model retains strong text performance while more effectively transferring those capabilities to audio inputs. The KL loss is computed on-policy based on the student's own generated trajectories, which we find effective in acquiring new speech capabilities and reducing forgetting of the backbone's original text knowledge~\citep{agarwal2024policy}.

\paragraph{Post-training with preference optimization.} To further improve the quality and usability of the model, we conduct post-training combining SFT on curated high-quality datasets, alongside a preference optimization approach. To improve computational and memory efficiency, we adopt SimPO~\citep{meng2024simpo}, a simplified preference optimization method that eliminates the need for a separate reference policy. SFT is applied to all tasks, while preference optimization is applied to STT, SpeechQA, SpokenQA, and TextQA. We train for 800 iterations with a global batch size of 96 and a packed sequence length of 8,192 tokens. We use a constant learning rate schedule with a peak learning rate of 1e-6 and a warmup of 112 steps.

We construct preference data using both offline and online approaches~\citep{li2025simplemix}. For offline data, we curate chosen-rejected prompt-response pairs specifically designed to discourage repetitive outputs. For online data, the chosen responses are taken from the SFT dataset, while the rejected responses are generated by our model given the same prompts, explicitly targeting and suppressing undesirable behaviors. To determine whether each generated response should be treated as a rejected sample, we evaluate it against the ground truth using task-specific reward functions. We use deterministic verifiers where applicable and LLM judge otherwise. If the judge rates a generated response higher than the ground truth, we exclude the pair from training. After preference optimization, we observe a consistent reduction in repetitive outputs and an improvement in reward scores across tasks.

\subsection{\RaonChat}

In this subsection, we describe how \RaonSpeech is extended into \RaonChat for real-time full-duplex interaction. The training process consists of three stages: causal \RaonSpeech\ adaptation, full-duplex pre-training, and full-duplex fine-tuning. We summarize the optimization setup associated with each stage.

\paragraph{Causal \RaonSpeech adaptation.} We first replace the original speech encoder from \RaonSpeech with a causal one and perform a short adaptation stage to obtain a causal \RaonSpeech\ initialization. This stage provides a stable starting point for the subsequent full-duplex training stages. We use the cross-entropy loss only during this stage. The speech encoder and speech codec both operate with their native sliding-window configurations, using chunk sizes of 30 seconds and 12 seconds, respectively. During training, we adopt a two-stage strategy, as \RaonSpeech, but with small training steps. In the alignment stage, we use AdamW with a peak learning rate of 1.5e-4, cosine decay to 1.5e-5, and a 700-step warmup. Training runs for 10,000 steps with a global batch size of 128, a maximum sequence length of 12,288 with sequence packing, and a 30-second audio limit. In the full training stage, we lower the learning rate to 2e-6, shorten the maximum sequence length to 8,192, increase the global batch size to 160, and expand the training data, while keeping the number of training steps unchanged.

\paragraph{Full-duplex pre-training.} After causal adaptation, we perform full-duplex pre-training on large-scale full-duplex interleaved data to expose the model to both the sequence patterns of simultaneous listening and speaking and a broad range of conversational behaviors. In this stage, the speech modules process full audio streams using their native sliding-window configurations rather than fixed training chunks, enabling training on continuous conversational dynamics in full-duplex settings. We also adopt one-frame text lookahead during this stage, so that speech semantic tokens are predicted conditioned on text generated one frame in advance. In addition, we broaden the training data to cover a wider range of real-time conversational patterns. To reduce forgetting and preserve the capabilities inherited from \RaonSpeech, we mix in 10\% of the original \RaonSpeech pre-training data. We use AdamW with a peak learning rate of 5e-6, cosine decay to 5e-7, and 5\% warmup. Training runs for 25,000 iterations with a global batch size of 128 and a maximum sequence length of 4,096 (approximately 2 minutes), using sequence packing. During training, we apply loss weights of 0.75 and 0.5 to \PAD and \SIL, respectively.

\paragraph{Full-duplex fine-tuning.} Finally, we conduct fine-tuning in two stages to progressively adapt the model to real-time conversational situations. In the first stage, the model is fine-tuned on a high-quality conversational dataset to establish stable full-duplex interaction patterns with accurate turn-taking,
natural response timing, and fluent speech generation. In the second stage, training shifts to synthetic and scenario-specific data covering a broader range of situations, including diverse persona-based dialogues, safety responses, self-repair patterns, user interruptions, and backchannel variations. We also provide the persona or dialogue context as a system prompt, following PersonaPlex~\citep{personaplex2026}. To stabilize the assistant persona across both stages, we fix the speaker identity during synthetic data generation. We further replace \EPAD with \BC for backchannel modeling, initialize the \BC embedding from the \EPAD embedding, and increase the weight of its cross-entropy loss by 50x to mitigate label imbalance. Each stage is trained for 5,000 iterations with AdamW, using a peak learning rate of 3e-6 and cosine decay to 3e-7, 5\% warmup, a global batch size of 64, and a maximum sequence length of 4,096 tokens with sequence packing. During fine-tuning, we set the loss weight for \SIL to 0.25.

\section{Data} \label{sec:data}

\begin{figure}[t]
    \centering    \includegraphics[width=\linewidth,height=0.6\textheight,keepaspectratio]{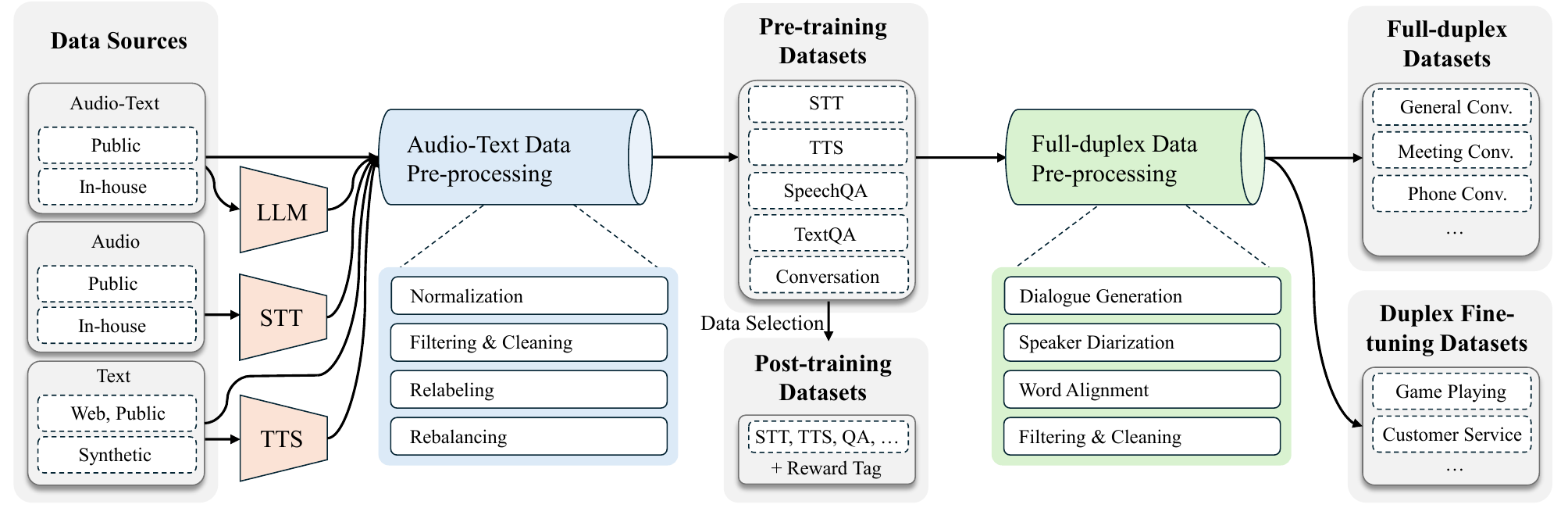}
    \caption{Overview of data curation. 
    (Left) data pre-processing pipeline for SpeechLMs. 
    (Right) data pre-processing pipeline for full-duplex models.}
    \label{fig:data_pipeline}
\end{figure}

Figure~\ref{fig:data_pipeline} illustrates an overview of the data curation pipeline for \RaonSpeech and \RaonChat. The training data is curated from multiple sources, comprising {1.38M} hours of English and Korean speech and text datasets.
With a comprehensive audio-text data curation pipeline, we construct high-quality speech pre-training and post-training datasets.
In addition, with a precise full-duplex data curation pipeline, we build full-duplex pre-training and fine-tuning datasets.
More detailed data statistics with token amounts are provided in \tabref{tab:data_statistics}.


\begin{table}[ht!]
\centering
\caption{Data statistics for \RaonSpeech and \RaonChat across training stages.}
\label{tab:data_statistics}
\resizebox{0.75\linewidth}{!}{%
\begin{tabular}{l c cc cc}
\toprule
& & \multicolumn{2}{c}{\textbf{\RaonSpeech}} & \multicolumn{2}{c}{\textbf{\RaonChat}} \\
\cmidrule(lr){3-4} \cmidrule(lr){5-6}
& Raw Data & Pre-training & Post-training & Pre-training & Fine-tuning \\
\midrule
Audio Hours  & 1.38M & 1.01M & 104.07K & 105.69K & 13.85K \\
\# Tokens    & 124.36B & 69.98B & 11.19B & 10.85B & 1.43B  \\
\# Samples   & 614.98M & 404.76M & 23.36M & 4.38M & 595.57K  \\
\bottomrule
\end{tabular}%
}
\end{table}

\subsection{SpeechLM Data Curation}

\paragraph{Data sources.} Constructing a large-scale audio-text dataset is challenging due to the limited amount of high-quality paired data. To address this, we leverage the abundance of audio-only and text-only data, in addition to audio-text paired data, to expand our dataset.
As these data are not paired, we generate the missing modality for each case using public STT models for audio data, and TTS models for text data.
In addition, to preserve original performance and further enhance reasoning capabilities, we include text-only datasets designed for text understanding tasks.
Specifically, each type of data is obtained:
\begin{itemize}[leftmargin=10pt,noitemsep]
  \item \textit{Audio-text}: Paired audio and text from public sources and in-house databases in English and Korean.
  Transcribed samples are curated into both STT and TTS formats through separate pipelines, while transcripts and audio metadata are further leveraged by LLMs to generate diverse task-specific data such as SpeechQA.
  \item \textit{Audio-only}: Speech data without transcriptions, collected from public corpora and web-sourced audio. Pseudo-label transcriptions are generated using STT models such as Whisper~\citep{radford2022whisper}, and the resulting speech-text pairs are incorporated into STT and SpeechQA tasks.
  \item \textit{Text-only}: Reading comprehension, commonsense reasoning, and instruction-following corpora.
  These serve directly as TextQA training data, or are converted into SpeechQA datasets by synthesizing questions or contexts into speech via TTS models, including Qwen3-TTS~\citep{hu2026qwen3tts}.
\end{itemize}

\paragraph{Data pre-processing.}
To improve the quality of the training data, we design a pre-processing method that filters, normalizes, relabels, and rebalances the dataset. The collected data exhibits quality issues from multiple sources: audio samples often contain severe noise, while text samples are overly informal or misaligned with realistic speech scenarios. These issues are further compounded in synthetic data, where transcriptions generated by STT models and speech synthesized by TTS models are not always accurate, and automatically generated transcripts can introduce additional noise. Beyond sample-level quality, the overall dataset distribution is biased or imbalanced in terms of topics or contents. To address these issues, we apply the following strategies:
\begin{itemize}[leftmargin=10pt,noitemsep]
  \item \textit{Normalization}: Apply primarily for STT and TTS tasks. Punctuation and capitalization are restored using a neural punctuation-and-capitalization model. Duplicate transcriptions, special symbols, and noise or repetition markers are further cleaned through rule-based normalization.
  \item \textit{Filtering}: Samples with speech-transcript mismatches, low-quality audio, low-quality QA pairs, or non-target languages are removed. For audio data, we apply STT-based error rate filtering, forced-alignment validation, and perceptual audio quality scoring. For LLM-generated outputs, we detect and remove repetitive or degenerate responses.
  \item \textit{Relabeling}: Low-quality transcripts are re-transcribed using Whisper, and QA answers are refined or regenerated using LLMs.
  \item \textit{Rebalancing}: The dataset composition is controlled along multiple axes, including audio domain, task type, and QA format, to ensure balanced capability coverage across all categories.
\end{itemize}

\subsection{Full-Duplex Data Curation}

\paragraph{Data sources.} Building full-duplex datasets is challenging due to the scarcity of time-aligned real-world conversational data at scale. To address this limitation, we train on a combination of real-world and synthetic duplex conversations. Real conversations provide naturally occurring interaction patterns and acoustic variability, while synthetic conversations offer scalable coverage of diverse scenarios and controllable placement of interactional events such as backchannels, overlap, and interruptions. Our synthetic data are generated with a dedicated pipeline that uses public LLMs and speech-based models to produce both transcripts and natural timing.

\begin{itemize}[leftmargin=10pt,noitemsep]
    \item \textit{Real Conversation}: 13.21K hours of conversational speech from diverse public corpora and in-house sources, capturing naturally occurring turn-taking, disfluencies, filler speech, and overlap under varied recording conditions.

    \item \textit{Synthetic Conversation}: 106.33K hours of conversational speech through a multi-stage pipeline. The data consists of (1) reconstructed samples from SpeechLM data to preserve the model’s prior knowledge, and (2) newly curated data tailored to target interaction scenarios. To improve robustness to abrupt topic shifts, we include context-free multi-turn dialogues and examples where the conversation returns to an earlier topic after intervening turns. The curated data covers diverse categories, including scenario-driven dialogues, open-domain chat, safety-critical interactions, and general-purpose conversations.

\end{itemize}

\paragraph{Data pre-processing.}

To construct full-duplex training data from both real and synthetic conversations, we employ a multi-stage data curation process. It consists of scenario-driven dialogue generation, speaker diarization, word alignment, and quality filtering, with each component selectively applied depending on the data type and sample characteristics.
This ensures high data fidelity and accurate temporal alignment.
Specifically, dialogue generation enables coverage of diverse interaction scenarios, while speaker diarization improves speaker-level separation. Word alignment enhances the temporal consistency between audio and text, and filtering further improves the overall data quality. The details of each stage is described below:

\begin{itemize}[leftmargin=10pt,noitemsep]
    \item \textit{Dialogue Generation}:
    For diverse interaction patterns across varying contexts, we design both task-oriented and speech-game scenarios, then generate dialogues using an LLM conditioned on them. The generated dialogues are then synthesized into speech using a TTS model. Each utterance is annotated with a conversational role, including \textit{speech}, \textit{backchannel}, \textit{interrupt}, and \textit{simultaneous speech}. We further refine the timing of interactional events such as backchannels, interruptions, and overlap using a combination of rule-based methods and a timing prediction model. Detailed implementations of the construction of duplex synthetic data are provided in the Appendix~\ref{app:fd_syc_pipeline}.
    \item \textit{Speaker Diarization}: As the majority of the real dataset is single-channel, we perform speaker diarization and extract target speech from overlapped regions.

    \item \textit{Word Alignment}:
    Time-aligned transcriptions are obtained using a forced-aligner model, with ground-truth input text when available and automatically generated transcriptions otherwise.
    \item \textit{Filtering}:
    Samples with misaligned backchannels or failed TTS synthesis are removed. We apply both rule-based and STT-based filtering to remove such samples, including those with excessively long audio relative to the transcript or with backchannels that do not overlap with the corresponding utterances. In addition, duplicated samples are identified via clustering and removed to reduce redundancy.
\end{itemize}

\section{Evaluation} \label{sec:evaluation}
We conduct a comprehensive evaluation of \RaonSpeech and \RaonChat against recent speech foundation models. \RaonSpeech is evaluated on both English and Korean benchmarks across automatic speech recognition (ASR), speech generation, spoken question answering, speech understanding, and text question answering, while \RaonChat is evaluated on English full-duplex dialogue benchmarks.

\subsection{\RaonSpeech}

\paragraph{Baselines.}
For \RaonSpeech, we compare against eight recent similarly sized
audio foundation models:
Qwen2.5-Omni~\citep{xu2025qwen3}, Kimi-Audio~\citep{ding2025kimi}, Audio Flamingo 3~\citep{goel2025audioflamingo}, Step-Audio 2 mini~\citep{wu2025step}, InteractiveOmni~\citep{tong2025interactiveomni}, Fun-Audio-Chat~\citep{team2025fun}, HyperCLOVA X 8B Omni~\citep{team2026hyperclova}, and MiniCPM-o 4.5~\citep{minicpmo45}.

\paragraph{Tasks and metrics.} We use four speech-related tasks and one text-related task to evaluate \RaonSpeech.
\begin{itemize}[leftmargin=12pt,itemsep=4pt,topsep=3pt,parsep=0pt]
    \item \textbf{Automatic Speech Recognition.} We evaluate four ASR benchmarks that cover both read and spontaneous speech under diverse acoustic conditions. For English, we use LibriSpeech~\citep{panayotov2015librispeech} and FLEURS~\citep{conneau2022fleurs}. For Korean, we evaluate on KSponSpeech~\citep{bang2020ksponspeech} and FLEURS. We report Word Error Rate (WER) for English and Character Error Rate (CER) for Korean to reflect language-specific characteristics. Lower values indicate better performance.

    \item \textbf{Speech Generation.} We assess speech generation quality on five benchmarks using both objective and perceptual metrics. For English, we use LibriSpeech test-clean and Seed~\citep{anastassiou2024seedtts}, and for Korean, we use KSponSpeech clean, MiniMax~\citep{zhang2025minimax}, and CV3-Eval~\citep{du2025cosyvoice}. We report WER for English and CER for Korean to measure intelligibility, computed by transcribing the generated speech using Whisper-large-v3~\citep{radford2022whisper} for English and a Zipformer~\citep{yao2023zipformer}-based in-house ASR model for Korean. We use UTMOSv2~\citep{baba2024utmosv2} to evaluate perceptual naturalness. Lower WER/CER and higher UTMOS indicate better performance.

    \item \textbf{Spoken Question Answering.} We evaluate spoken question answering capabilities using audio questions as input.
    For English, we employ VoiceBench~\citep{chen2024voicebench} and OpenAudioBench, introduced together with Baichuan-Audio~\citep{li2025openaudiobench}. VoiceBench comprises OpenBookQA, MMSU, Big-Bench-Hard~(BBH), SD-QA, AlpacaEval, CommonEval, WildVoice, IFEval, and AdvBench, while OpenAudioBench consists of AlpacaEval, Llama Questions (LlamaQ), TriviaQA, and Web Questions (WebQ).

    As no benchmark exists for the Korean spoken question answering task, we construct new benchmarks, KVoiceBench\footnote{\url{https://huggingface.co/datasets/KRAFTON/KVoiceBench}} and KOpenAudioBench\footnote{\url{https://huggingface.co/datasets/KRAFTON/KOpenAudioBench}}. Specifically, we translate all transcriptions from VoiceBench and OpenAudioBench into Korean, normalize them into speech-friendly text, and synthesize them using a Qwen3-TTS system. During this process, we remove or adapt non-transferable linguistic features (e.g., capitalization and certain grammatical rules) to better align with Korean.

    For evaluation, we report accuracy for multiple-choice questions (OpenBookQA, MMSU, BBH) and short-answer questions (LlamaQ, TriviaQA, WebQ).
    For open-ended questions (AlpacaEval, CommonEval, and WildVoice), we use GPT-5.4~\citep{singh2025openai} as a judge and report the scores on a 100-point scale. We adopt the judge prompt provided by VoiceBench for English, and use its translated version for Korean.
    For IFEval, we report the average of prompt-level accuracy and instruction-level accuracy, each computed as the average of strict and loose scores. For AdvBench, we report the refusal rate using rule-based phrase detection. For readability, the main English and Korean result tables report aggregate VoiceBench/OpenAudioBench and KVoiceBench/KOpenAudioBench scores, while Appendix~\ref{app:spokenqa_details} provides the per-benchmark spoken question answering breakdowns.

    \item \textbf{Speech Understanding.} We use the speech subset of MMAU (test-mini split)~\citep{sakshi2024mmau} and MMAU-Pro~\citep{kumar2025mmaupro} for English. Since no prior benchmark is designed for the Korean speech understanding, we construct a new benchmark, KMMAU~\footnote{\url{https://huggingface.co/datasets/KRAFTON/KMMAU}}. KMMAU is built using audio, metadata, and transcriptions from three Korean audio datasets: KSS~\citep{kss}, KMSAV~\citep{kmsav}, and Seoul Corpus~\citep{SeoulCorpus}.
    From these sources, we derive capability-level questions covering speaker counting, speaker-attribute recognition such as gender and age, fact extraction, topic understanding, and word-level reasoning using the associated audio, metadata, and transcriptions. In the condensed summary table, KMMAU is reported as the average of the capability-level accuracies shown in Appendix~\ref{app:audio_understanding_groups}. These detailed capability-level results clarify which aspects of Korean speech understanding are already strong and which remain difficult. As all speech understanding benchmarks are formulated as multiple-choice questions, we report accuracy as the evaluation metric.

    \item \textbf{Text Question Answering.} To examine whether training on the speech modality induces catastrophic forgetting in the backbone LLM, we additionally evaluate performance on text question answering tasks. For English, we use MMLU-Pro~\citep{wang2024mmlupro} and MMLU-Redux~\citep{gema2025mmluredux}, and for Korean, we evaluate KMMLU-Pro and KMMLU-Redux~\citep{hong2025kmmlureduxkmmlupro}. All results are reported in terms of accuracy.

\end{itemize}

\definecolor{lightgray}{gray}{0.90}
\providecommand{\best}[1]{\textbf{#1}}
\providecommand{\secondbest}[1]{\underline{#1}}

\begin{table*}[t]
\centering\caption{English speech and text benchmark results for \RaonSpeech. \best{Bold} and \secondbest{underline} indicate the best and the second-best performance, respectively.}
\label{tab:eng_speechlm_performance}

\small
\resizebox{\textwidth}{!}{
\begin{tabular}{l>{\columncolor{blue!10}}c cccccccc}
\toprule
\textbf{Benchmark} & \textbf{Raon} & \textbf{Qwen2.5} & \textbf{Kimi} & \textbf{Audio} & \textbf{Step-Audio} & \textbf{Interactive} & \textbf{Fun-Audio} & \textbf{HyperCLOVA} & \textbf{MiniCPM} \\
 & \textbf{-Speech} & \textbf{-Omni} & \textbf{-Audio} & \textbf{Flamingo 3} & \textbf{2 mini} & \textbf{Omni} & \textbf{Chat} & \textbf{X 8B Omni} & \textbf{-o 4.5}\\

\midrule
\rowcolor{lightgray}
\multicolumn{10}{c}{\textbf{Automatic Speech Recognition (WER $\downarrow$)}} \\
\midrule

LibriSpeech-c & 1.44 & 1.73 & \best{1.38} & \secondbest{1.40} & 4.88 & 2.28 & 1.60 & 2.28 & 1.51 \\
LibriSpeech-o & \secondbest{2.89} & 3.88 & \best{2.70} & 2.97 & 6.82 & 4.67 & 3.89 & 5.03 & 3.56 \\
Fleurs-en & \secondbest{3.59} & 4.05 & 4.54 & 4.54 & 13.02 & 4.89 & 7.61 & 5.57 & \best{3.52} \\

\midrule
\rowcolor{lightgray}
\multicolumn{10}{c}{\textbf{Speech Generation (WER $\downarrow$ | UTMOS $\uparrow$)}} \\
\midrule

LibriSpeech-c & \best{2.01} | 3.26 & 2.30 | 3.55 & -- & -- & 3.01 | \best{3.83} & 3.11 | \secondbest{3.68} & 72.52 | 3.33 & 7.31 | 3.23 & 11.08 | 3.37 \\
Seed & \best{1.93} | 3.20 & 3.54 | 3.56 & -- & -- & 3.49 | \best{3.85} & 2.70 | \secondbest{3.69} & 22.26 | 3.38 & 3.42 | 3.29 & 4.72 | 3.06 \\

\midrule
\rowcolor{lightgray}
\multicolumn{10}{c}{\textbf{Spoken Question Answering} $\uparrow$} \\
\midrule

VoiceBench & \best{76.79} & 66.71 & 68.92 & 41.60 & 50.26 & 62.41 & 73.64 & 48.70 & \secondbest{76.06} \\
OpenAudioBench & 70.21 & 66.73 & 68.23 & 38.88 & 59.63 & 66.68 & \secondbest{72.39} & 57.44 & \best{74.82} \\

\midrule
\rowcolor{lightgray}
\multicolumn{10}{c}{\textbf{Speech Understanding (Accuracy $\uparrow$)}} \\
\midrule

MMAU (Speech) & \best{78.68} & \secondbest{77.18} & 66.37 & 68.77 & 68.47 & 66.07 & 71.47 & 53.15 & 72.67 \\
MMAU-Pro (Speech)  & \best{64.65} & 62.74 & 54.77 & 52.41 & 59.60 & 44.11 & \secondbest{64.53} & 40.52 & 59.48 \\

\midrule
\rowcolor{lightgray}
\multicolumn{10}{c}{\textbf{Text Question Answering (Accuracy $\uparrow$)}} \\
\midrule

MMLU-Pro & \best{64.05} & 50.40 & 16.66 & 2.52 & 34.95 & 31.38 & \secondbest{61.12} & 53.79 & 55.20 \\
MMLU-Redux  & \best{78.87} & 68.03 & 44.27 & 0.90 & 51.73 & 36.03 & \secondbest{74.70} & 71.83 & 72.53 \\

\bottomrule
\end{tabular}
}
\end{table*}

\definecolor{lightgray}{gray}{0.90}
\providecommand{\best}[1]{\textbf{#1}}
\providecommand{\secondbest}[1]{\underline{#1}}

\begin{table*}[t]
\centering
\caption{Korean speech and text benchmark results for \RaonSpeech. \best{Bold} and \secondbest{underline} indicate the best and the second-best performance, respectively.}
\label{tab:kor_speechlm_performance}
\small
\resizebox{\textwidth}{!}{
\begin{tabular}{l>{\columncolor{blue!10}}c ccccccc}
\toprule
\textbf{Benchmark} & \textbf{Raon} & \textbf{Qwen2.5} & \textbf{Audio} & \textbf{Step-Audio} & \textbf{Interactive} & \textbf{Fun-Audio} & \textbf{HyperCLOVA} & \textbf{MiniCPM} \\
 & \textbf{-Speech} & \textbf{-Omni} & \textbf{Flamingo 3} & \textbf{2 mini} & \textbf{Omni} & \textbf{Chat} & \textbf{X 8B Omni} & \textbf{-o 4.5}\\

\midrule
\rowcolor{lightgray}
\multicolumn{9}{c}{\textbf{Automatic Speech Recognition (CER $\downarrow$)}} \\
\midrule

KSponSpeech-c & \best{6.56} & 18.96 & 134.12 & 55.84 & 461.87 & 646.25 & \secondbest{10.22} & 205.35 \\
KSponSpeech-o & \best{6.96} & 22.72 & 136.50 & 59.43 & 428.83 & 514.82 & \secondbest{10.15} & 202.14 \\
Fleurs-ko & \best{1.81} & \secondbest{3.24} & 71.85 & 45.72 & 159.10 & 36.44 & 3.70 & 168.14 \\

\midrule
\rowcolor{lightgray}
\multicolumn{9}{c}{\textbf{Speech Generation (CER $\downarrow$ | UTMOS $\uparrow$)}} \\
\midrule

KSponSpeech-c & \best{4.89} | 2.36 & 121 | 2.82 & -- & 28.13 | \best{3.27} & 98.93 | \secondbest{3.10} & 112.06 | 2.95 & \secondbest{16.7} | 2.71 & 111.02 | 2.77 \\
MiniMax-ko & \best{1.57} | 2.88 & 121 | 2.92 & -- & 23.35 | \best{3.54} & 99.88 | 3.12 & 70.60 | 3.00 & \secondbest{2.64} | \secondbest{3.24} & 103.69 | 2.71 \\
CV3-Eval-ko & \best{3.90} | 2.64 & 118 | 2.96 & -- & 35.33 | \best{3.46} & 96.12 | 3.20 & 85.72 | 2.97 & \secondbest{4.52} | \secondbest{3.29} & 117.46 | 2.68 \\

\midrule
\rowcolor{lightgray}
\multicolumn{9}{c}{\textbf{Spoken Question Answering} $\uparrow$} \\
\midrule

KVoiceBench & \best{66.62} & 49.04 & 18.82 & 32.03 & 19.96 & \secondbest{50.12} & 45.11 & 39.47 \\
KOpenAudioBench & \best{52.10} & 39.23 & 12.60 & 31.00 & 11.45 & 43.05 & \secondbest{45.09} & 35.66 \\

\midrule
\rowcolor{lightgray}
\multicolumn{9}{c}{\textbf{Speech Understanding (Accuracy $\uparrow$)}} \\
\midrule

KMMAU & \best{71.83} & 62.85 & 44.46 & 63.02 & 30.56 & \secondbest{67.37} & 30.99 & 62.39 \\

\midrule
\rowcolor{lightgray}
\multicolumn{9}{c}{\textbf{Text Question Answering (Accuracy $\uparrow$)}} \\
\midrule

KMMLU-Pro & \best{46.85} & 32.49 & 0.43 & 38.38 & 36.43 & \secondbest{43.23} & 19.06 & 41.57 \\
KMMLU-Redux  & \best{51.80} & 30.54 & 0.27 & 35.41 & 34.98 & 45.07 & 30.58 & \secondbest{46.27} \\

\bottomrule
\end{tabular}
}
\end{table*}

\paragraph{Results.} Tables~\ref{tab:eng_speechlm_performance} and~\ref{tab:kor_speechlm_performance} report the English and Korean benchmark results for \RaonSpeech, and Appendix~\ref{app:spokenqa_details} reports the per-benchmark spoken question answering breakdowns underlying the aggregate suite rows. Overall, \RaonSpeech shows the strongest speech-centric profile in our main comparison. In English, its clearest gains are in speech understanding, spoken question answering, and generated-speech intelligibility: it achieves the best scores on MMAU and MMAU-Pro, the highest average score on VoiceBench, and the lowest WER on English speech-generation benchmarks, while remaining competitive on OpenAudioBench and ASR. These gains are not achieved at the expense of text capability, as \RaonSpeech also achieves the best MMLU-Pro and MMLU-Redux results. In Korean, the gains are broader and stronger. \RaonSpeech achieves the best CER on all three ASR benchmarks and all three speech-generation benchmarks, together with the best KVoiceBench, KOpenAudioBench, KMMAU, KMMLU-Pro, and KMMLU-Redux results. The appendix breakdown further shows that it leads 10 of the 12 Korean spoken question answering benchmarks. Taken together, the results suggest that the largest gains come from speech-centric capabilities, especially Korean speech perception, speech understanding, and generated-speech intelligibility, while perceptual naturalness remains a relative strength of some baselines as reflected by UTMOS.

\providecommand{\fdpending}{\textcolor{BrickRed}{TBD}}
\providecommand{\fdpendingtext}[1]{\textcolor{BrickRed}{#1}}

\subsection{\RaonChat}

\paragraph{Baselines.}
We compare \RaonChat with Moshi~\citep{defossez2024moshi}, Freeze-Omni~\citep{freezeomni2024}, PersonaPlex~\citep{personaplex2026}, and MiniCPM-o 4.5~\citep{minicpmo45}.

\paragraph{Full-duplex speech dialogue.}
For \RaonChat, we evaluate on Full-Duplex-Bench (FDB) v1.0, v1.5, and v2.0~\citep{fdbv1,fdbv15,fdbv2}. For FDB v1.0 and v1.5, we use the official benchmark sets and report scores reproduced with an internal evaluator. Appendix~\ref{app:fd_eval_details} summarizes the main differences between our offline evaluator and the public FDB v1.0/v1.5 reference scripts.

\begin{itemize}[leftmargin=12pt,itemsep=4pt,topsep=3pt,parsep=0pt]
    \item \textbf{FDB v1.0.} FDB v1.0 evaluates four core turn-taking behaviors using prerecorded conversations. Pause handling assesses the model's ability to avoid taking the floor during short within-speaker pauses. Backchanneling tests the model's ability to produce brief acknowledgements at appropriate times and frequencies without taking the floor. Smooth turn-taking measures how naturally the model takes the floor after the speaker yields it, and user interruption tests whether the model stops and responds appropriately when the user barges in. The benchmark uses CANDOR~\citep{reece2023candor} for pause handling and smooth turn-taking, the In-Conversation Corpus (ICC)~\citep{umair2024speak} for backchanneling, and synthetic data for controlled pause-handling and interruption scenarios.

    \item \textbf{FDB v1.5.} FDB v1.5 keeps the offline protocol but focuses on overlapped speech. User Backchannel evaluates the model's ability to continue its response when the user produces a short acknowledgement. Background Speech assesses whether the model ignores irrelevant ambient speech while continuing the answer. Talking to Others tests whether the model ignores speech that is not addressed to it and stays on the ongoing interaction. User Interruption requires the model to stop and respond to new speech addressed to it. In the main text, we report the scenario-wise Resume or Respond rate together with the Unknown rate, and defer the latency details to Appendix~\ref{app:fd_eval_details}.

    \item \textbf{FDB v2.0.} FDB v2.0 is a multi-turn evaluation framework with an automated examiner, consisting of four task types: Daily, Correction, Entity Tracking, and Safety. Daily evaluates routine open-domain conversations, Correction evaluates whether the model can revise or repair its response after follow-up feedback, Entity Tracking evaluates whether the model maintains and updates dialogue state across turns, and Safety evaluates whether the model follows safety-critical conversational constraints. The benchmark reports Turn-Taking Fluency, Multi-Turn Instruction Following, and a Task-Specific Metric, where the last score summarizes task completion within each task family.
\end{itemize}

\paragraph{Metrics.}
We report metrics across FDB versions to evaluate turn-taking behavior, backchanneling, and task performance.

\begin{itemize}[leftmargin=10pt,itemsep=4pt,topsep=2pt]
  \item \textbf{FDB v1.0.} Takeover Rate (TOR) quantifies how often takeovers occur during the conversation. Lower TOR is preferred for pause handling and backchanneling, whereas higher TOR is preferred for smooth turn-taking and user interruption. Backchannel Frequency (Freq.) reflects how often a model produces backchannels without taking the turn. Latency denotes the average response time after an interruption or the end of user speech. Jensen-Shannon Divergence (JSD) measures the difference between the model's predicted backchannel timing distribution and human timing. Judge denotes an LLM-based score measuring the relevance of responses to user interruptions.
  \item \textbf{FDB v1.5.} Resume is reported for User Backchannel, Background Speech, and Talking to Others, while Respond is reported for User Interruption. Unknown is also included, as the official behavior annotation requires a valid post-overlap segment.
  \item \textbf{FDB v2.0.} Turn-Taking Fluency (TT Fluency) measures how natural and well-timed responses are during turn-taking. Multi-Turn Instruction Following (IF) measures how well the model interprets and executes instructions across turns. Task-Specific Metric (Task Metric) evaluates overall task completion.
\end{itemize}

\paragraph{Results.}

Table~\ref{tab:fd_summary} reports the main results of FDB v1.0, v1.5, and v2.0 for \RaonChat.
Appendix~\ref{app:fd_eval_details} provides scenario-wise FDB v1.5 latency details, the full FDB v1.5 behavior distributions, and the main differences in the offline evaluator.

Overall, \RaonChat shows its clearest gains on interruption-sensitive turn-taking and overlap-robust response behavior. On FDB v1.0, it demonstrates leading performance across the majority of metrics, achieving the best backchannel TOR and frequency, the best user-interruption TOR, and near-best pause handling and smooth turn-taking. On FDB v1.5, it remains competitive on Background Speech, Talking to Others, and User Interruption, including the best Unknown rate on User Interruption and second-best Unknown rates on Background Speech and Talking to Others, although performance on User Backchannel remains weaker than that of the strongest baselines. On FDB v2.0, \RaonChat improves over Moshi and Freeze-Omni on all three session-level metrics, while PersonaPlex and MiniCPM-o 4.5 remain stronger on this long-horizon multi-turn setting.

\providecommand{\fdpending}{\textcolor{BrickRed}{TBD}}
\providecommand{\fdpendingtext}[1]{\textcolor{BrickRed}{#1}}
\providecommand{\fdbest}[1]{\textbf{#1}}
\providecommand{\fdsecondbest}[1]{\underline{#1}}
\providecommand{\fdUp}{(\textbf{$\uparrow$})}
\providecommand{\fdDown}{(\textbf{$\downarrow$})}
\definecolor{lightgray}{gray}{0.90}

\begin{table*}[t]
\centering
\caption{Full-Duplex-Bench summary for Raon-SpeechChat across FDB v1.0, v1.5, and v2.0. \best{Bold} and \secondbest{underline} indicate the best and the second-best performance, respectively.}
\label{tab:fd_summary}
\footnotesize
\renewcommand{\arraystretch}{1.10}
\setlength{\tabcolsep}{2.5pt}

\begin{tabular*}{0.98\textwidth}{@{\extracolsep{\fill}}
                >{\raggedright\arraybackslash}m{2.42cm}
                >{\raggedright\arraybackslash}m{2.42cm}
                >{\centering\arraybackslash\columncolor{blue!10}}m{1.55cm}
                *{4}{>{\centering\arraybackslash}m{1.43cm}}}
\toprule
\multicolumn{2}{c}{\textbf{Benchmark Slice}} & \multicolumn{5}{c}{\textbf{Models}} \\
\cmidrule{1-2} \cmidrule{3-7}
\textbf{\makecell[l]{Scenario / Task}} & \textbf{Metric} & \textbf{\makecell[c]{Raon-\\SpeechChat}} & \textbf{Moshi} & \textbf{\makecell[c]{Freeze-\\Omni}} & \textbf{\makecell[c]{Persona\\Plex}} & \textbf{\makecell[c]{MiniCPM-o\\4.5}} \\
\midrule
\rowcolor{lightgray}
\multicolumn{7}{c}{\textbf{FDB v1.0}} \\
\cmidrule{1-7}
\multirow{2}{*}{\makecell[l]{Pause\\Handling}} & Synthetic TOR \fdDown & \fdsecondbest{0.212} & 0.299 & 0.620 & \fdsecondbest{0.212} & \fdbest{0.182} \\
 & Candor TOR \fdDown & \fdsecondbest{0.213} & 0.370 & 0.435 & \fdbest{0.204} & 0.343 \\
\cmidrule{1-7}
\multirow{3}{*}{Backchannel} & TOR \fdDown & \fdbest{0.091} & 0.309 & 0.564 & \fdsecondbest{0.236} & 0.418 \\
 & Freq. \fdUp & \fdbest{0.081} & \fdsecondbest{0.050} & 0.002 & 0.046 & 0.008 \\
 & JSD \fdDown & 0.775 & \fdbest{0.664} & 0.983 & \fdsecondbest{0.735} & 0.899 \\
\cmidrule{1-7}
\multirow{2}{*}{\makecell[l]{Smooth\\Turn-Taking}} & TOR \fdUp & \fdsecondbest{0.832} & 0.437 & 0.252 & 0.782 & \fdbest{0.891} \\
 & Latency \fdDown & 1.034 & \fdbest{0.726} & 1.136 & 1.101 & \fdsecondbest{1.000} \\
\cmidrule{1-7}
\multirow{3}{*}{\makecell[l]{User\\Interruption}} & Judge \fdUp & 2.790 & 2.908 & 2.830 & \fdsecondbest{2.943} & \fdbest{3.408} \\
 & TOR \fdUp & \fdbest{0.980} & 0.705 & \fdsecondbest{0.910} & 0.880 & 0.845 \\
 & Latency \fdDown & \fdsecondbest{1.219} & 2.027 & 2.270 & \fdbest{0.980} & 1.233 \\
\midrule
\rowcolor{lightgray}
\multicolumn{7}{c}{\textbf{FDB v1.5}} \\
\cmidrule{1-7}
\multirow{2}{*}{\makecell[l]{User\\Backchannel}} & Resume \fdUp & 0.398 & 0.092 & \fdsecondbest{0.480} & 0.418 & \fdbest{0.520} \\
 & Unknown \fdDown & 0.582 & 0.898 & \fdsecondbest{0.490} & 0.520 & \fdbest{0.480} \\
\cmidrule{1-7}
\multirow{2}{*}{\makecell[l]{Background\\Speech}} & Resume \fdUp & \fdsecondbest{0.230} & 0.100 & 0.100 & 0.160 & \fdbest{0.260} \\
 & Unknown \fdDown & \fdsecondbest{0.220} & 0.660 & \fdbest{0.130} & 0.510 & 0.280 \\
\cmidrule{1-7}
\multirow{2}{*}{\makecell[l]{Talking to\\Others}} & Resume \fdUp & \fdsecondbest{0.150} & \fdbest{0.210} & \fdsecondbest{0.150} & 0.120 & 0.130 \\
 & Unknown \fdDown & \fdsecondbest{0.190} & 0.550 & \fdbest{0.180} & 0.370 & 0.310 \\
\cmidrule{1-7}
\multirow{2}{*}{\makecell[l]{User\\Interruption}} & Respond \fdUp & \fdsecondbest{0.725} & 0.560 & \fdbest{0.810} & 0.710 & 0.660 \\
 & Unknown \fdDown & \fdbest{0.085} & 0.275 & \fdsecondbest{0.090} & 0.115 & 0.115 \\
\midrule
\rowcolor{lightgray}
\multicolumn{7}{c}{\textbf{FDB v2.0}} \\
\cmidrule{1-7}
\multirow{3}{*}{\makecell[l]{Multi-Turn\\Session}} & TT Fluency \fdUp & 3.552 & 3.274 & 3.176 & \fdsecondbest{3.706} & \fdbest{3.984} \\
 & IF \fdUp & 3.042 & 2.533 & 2.610 & \fdsecondbest{3.162} & \fdbest{3.534} \\
 & Task Metric \fdUp & 2.944 & 2.259 & 2.426 & \fdsecondbest{3.111} & \fdbest{3.241} \\
\bottomrule
\end{tabular*}
\end{table*}

\section{Related Work} \label{sec:related_work}
\paragraph{Speech language models.}
Recent years have seen rapid advancements in SpeechLMs, marked by three major trends: the transition from speech understanding to joint understanding and generation, the expansion toward omni-modal modeling across image, video, audio, and text, and the development of real-time interactive capabilities and broader language coverage. Qwen2-Audio~\citep{chu2024qwen2audio} is an early representative model that jointly trains on diverse audio tasks to support both speech understanding and generation.
This direction has further evolved into omni-modal models such as Qwen3-Omni~\citep{xu2025qwen3}, MiniCPM-o 4.5~\citep{minicpmo45}, and InteractiveOmni~\citep{tong2025interactiveomni}, which extend multimodal reasoning and dialogue beyond audio and text to include image and video.
For audio-centric and interaction quality, models such as Kimi-Audio~\citep{ding2025kimi} and Audio Flamingo 3~\citep{goel2025audioflamingo} achieve strong performance through large-scale audio-text pretraining, while Step-Audio 2 mini~\citep{wu2025step} improves efficient speech comprehension and Fun-Audio-Chat~\citep{team2025fun} focuses on natural multi-turn spoken interaction. Despite these advances, most existing models remain centered on high-resource languages such as English and Chinese, limiting their effectiveness in broader linguistic settings. Although recent efforts such as HyperCLOVA X Omni~\citep{team2026hyperclova} support additional languages including Korean, substantial performance gaps still remain.

\paragraph{Full-duplex models.}
To support more natural and realistic spoken interactions, full-duplex models have been proposed beyond the explicit turn-taking assumption of conventional SpeechLMs by enabling simultaneous listening and speaking.
Moshi~\citep{defossez2024moshi} is a pioneering full-duplex speech-text foundation model that introduces an inner monologue mechanism, enabling real-time spoken dialogue without explicit turn-taking signals.
Freeze-Omni~\citep{freezeomni2024} proposes a low-latency speech-to-speech dialogue framework that keeps the LLM backbone frozen, enabling efficient adaptation for real-time interaction.
OmniFlatten~\citep{zhang2025omniflatten} introduces a progressive training scheme to flatten text-based LLMs into full-duplex speech models while preserving language understanding capability.
PersonaPlex~\citep{personaplex2026} builds on the Moshi architecture and introduces voice and role control mechanisms, enabling consistent persona-aware interaction.
While these models demonstrate promising full-duplex capabilities, they still show limitations in delicate temporal awareness and naturalness.

\section{Discussion} \label{sec:discussion}
\paragraph{Conclusion.}
We present \textbf{\RaonSpeech} and \textbf{\RaonChat}, a strong bilingual SpeechLM and its full-duplex extension for natural real-time conversation.
\RaonSpeech successfully transforms a pretrained LLM into a high-quality SpeechLM through a three-stage training pipeline, establishing the strongest speech-centric profile in our benchmark suite across 42 English and Korean speech and text benchmarks against eight similarly sized recent audio foundation models while preserving strong text capabilities.
\RaonChat further enables natural full-duplex spoken interaction through continual training on large-scale time-aligned conversational data, showing its clearest strengths on the turn-taking and interruption-sensitive behaviors covered by FDB v1.0 while remaining competitive across the broader full-duplex evaluation suite. Long-horizon multi-turn instruction following remains an important target for future work.
As we open-source all model checkpoints, inference code, and an interactive demo, we believe that our collection will have a far-reaching impact on accessible and practical speech-language interaction research.

\paragraph{Future work.}
While \RaonSpeech and \RaonChat demonstrate impressive performance, several directions remain for future work.
First, beyond bilingual settings, we can extend our framework into more diverse languages to support truly multilingual SpeechLMs.
Second, integrating vision modality into \RaonSpeech and \RaonChat would enable richer multimodal interaction, allowing the model to jointly reason over speech, audio, and visual inputs in real-time.
Finally, we can extend our models toward agentic tasks by post-training on speech-natured environments, which would lead to speech-driven agents capable of stably executing complex, multi-step tasks through spoken interaction.

\section{Authorship and Credit Assignment}

\vspace{0.1cm}

Within each role, names are alphabetically arranged by first name, then by last name. The leads of each role are marked by $^*$.

\begin{center}
{\large \textbf{Core Contributors}}
\end{center}

\paragraph{Modeling.} Ethan Ewer, Gyeongman Kim, Jihun Yun, Joonghyun Bae, Junhyuck Kim, Sehun Lee, and Keon Lee$^*$.

\paragraph{Data.} Beomsoo Kim, Changho Choi, Dohyun Kim, Eunchong Kim, Minkyu Kim, Sungwoo Cho, and Dongmin Park$^*$.

\paragraph{Evaluation.} Haechan Kim, Inkyu Park, Seungjun Chung, and Jonghyun Lee$^*$.

\paragraph{Serving and engineering.} Hyeonghwan Kim, Jihwan Moon, and Dongwon Kim$^*$.

\paragraph{Infrastructure.} Jiyun Kim, Dongki Lee, and Hara Kang$^*$.

\paragraph{Project leaders.} Kangwook Lee, and Jaewoong Cho$^*$.

\begin{center}
{\large \textbf{Acknowledgements}}
\end{center}

\noindent We would like to express our sincere gratitude to the following individuals for their valuable contributions and support to this work.

\noindent Beongjun Choi, Howon Lee, Hyeonah Park, Jaeyun Song, Jihoo Lee, Jinwoo Kim, Junhyoung Chung, Junkyu Park, Sihyeong Park, and Taehong Moon.

\bibliography{reference}

\clearpage
\appendix

\section{Performance on Korean Speech Benchmarks}\label{app:kor_overall_performance}

\begin{figure}[h]
    \centering    \includegraphics[width=0.6\linewidth,height=0.6\textheight,keepaspectratio]{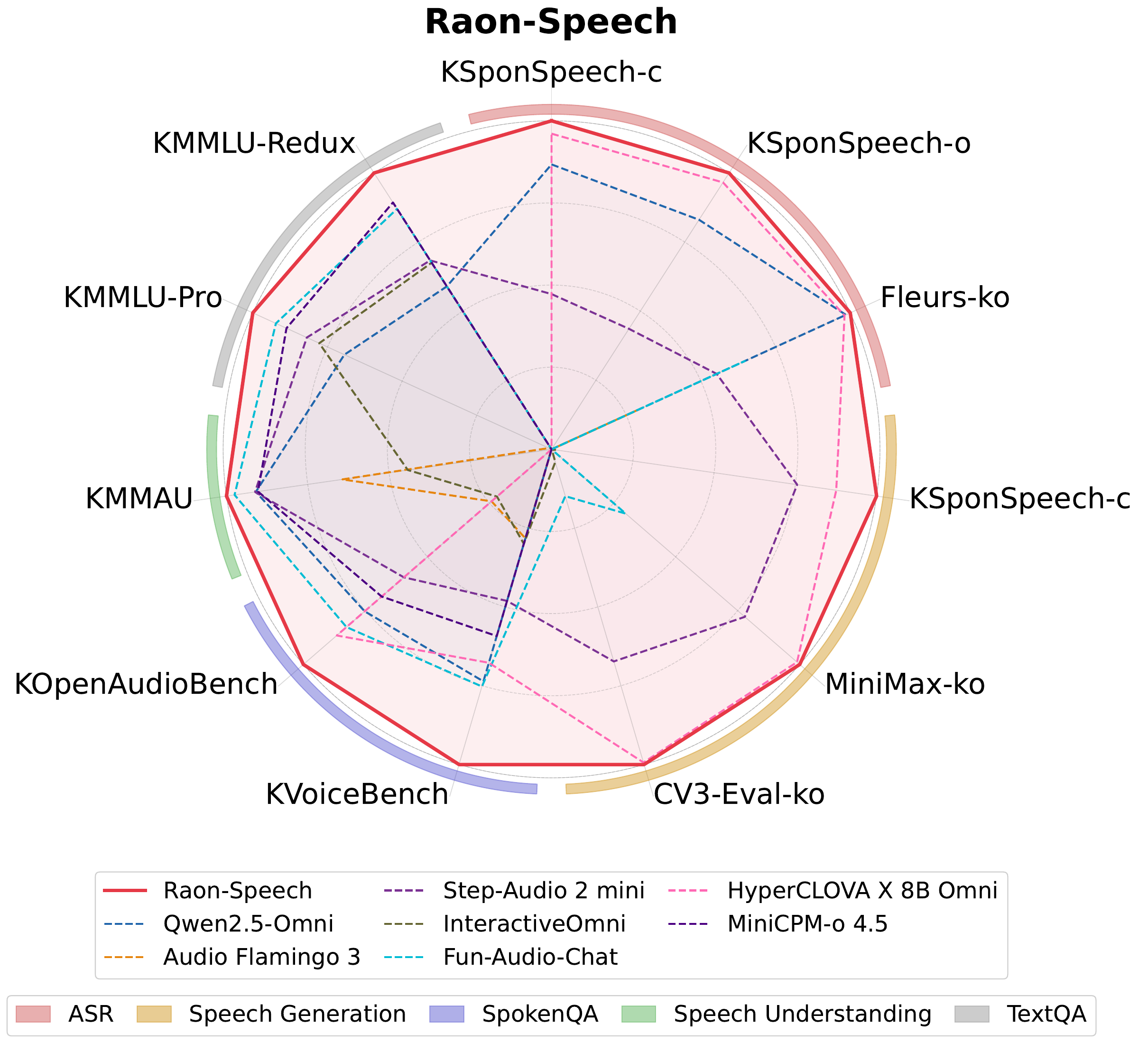}
    \caption{Overall performance comparison of Raon-Speech against baseline models on Korean speech and text benchmarks spanning automatic speech recognition (ASR), speech generation, spoken question answering (SpokenQA), speech understanding,
   and text question answering (TextQA). All scores are zero to max normalized per benchmark axis.}
    \label{fig:main_figure_kor}
\end{figure}

Figure~\ref{fig:main_figure_kor} shows the overall performance of \RaonSpeech on diverse Korean speech and text benchmarks.

\section{Module-Wise Parameter Breakdown}\label{app:param_breakdown}

Table~\ref{tab:param_counts} provides a module-wise breakdown of parameter counts for \RaonSpeech and \RaonChat.

\begin{table*}[hbpt]
\centering
\footnotesize
\setlength{\tabcolsep}{6pt}
\renewcommand{\arraystretch}{1.08}
\caption{Module-wise parameter counts of \RaonSpeech and \RaonChat. Modules above one billion parameters are reported in billions (B), and smaller modules are reported in rounded millions (M). Speaker conditioning includes both the runtime-loaded ECAPA-TDNN backbone and the learned projection.}
\label{tab:param_counts}
\begin{tabular}{@{}l r r@{}}
\toprule
\textbf{Module} & \textbf{\RaonSpeech} & \textbf{\RaonChat} \\
\midrule

\multicolumn{3}{c}{\textbf{Shared modules}} \\
\midrule 
LLM Backbone (incl.\ LM Head) & \multicolumn{2}{c}{8.2B} \\
Speech Generation Expert (incl.\ Audio LM Head) & \multicolumn{2}{c}{205M} \\
LLM-to-Speech-Generation-Expert Projector & \multicolumn{2}{c}{38M} \\
RCP (incl.\ input projector) & \multicolumn{2}{c}{146M} \\
Speech Codec & \multicolumn{2}{c}{96M} \\
Speech Output Adaptor & \multicolumn{2}{c}{19M} \\
Speaker Conditioning & \multicolumn{2}{c}{15M} \\
\midrule 
\textit{Shared modules subtotal} & \multicolumn{2}{c}{\textit{8.8B}} \\
\midrule

\multicolumn{3}{c}{\textbf{Variant-specific modules}} \\
\midrule 
Speech Encoder & 319M & 970M \\
Speech Input Adaptor & 25M & 38M \\
\midrule 
\textit{Variant-specific subtotal} & \textit{344M} & \textit{1,008M} \\
\midrule

\textbf{Full model} & \textbf{9.1B} & \textbf{9.8B} \\
\bottomrule
\end{tabular}
\end{table*}

\section{Detailed Model Configuration}\label{app:model_config}


\newcommand{\rowsep}{\cmidrule(l{6pt}r{6pt}){1-3}}

\begin{table*}[t]
\centering
\footnotesize
\setlength{\tabcolsep}{4.5pt}
\renewcommand{\arraystretch}{1.10}
\caption{Detailed architectural comparison of \RaonSpeech and \RaonChat, with shared modules listed once and speech encoder-related components compared separately. SWA denotes sliding-window attention.}
\label{tab:arch_compact}
\begin{tabular}{@{}
    >{\raggedright\arraybackslash}p{0.20\textwidth}
    >{\raggedright\arraybackslash}p{0.38\textwidth}
    >{\raggedright\arraybackslash}p{0.38\textwidth}
@{}}
\toprule
\multicolumn{1}{c}{\textbf{Component}} &
\multicolumn{1}{c}{\textbf{\RaonSpeech}} &
\multicolumn{1}{c}{\textbf{\RaonChat}} \\
\midrule

\multicolumn{3}{c}{\textbf{Shared modules}} \\
\midrule

LLM Backbone &
\multicolumn{2}{p{0.76\textwidth}@{}}{\raggedright
Hidden 4096, 36 layers, FFN 12288 (SwiGLU), RoPE
} \\
\midrule 

Speech Generation Stack &
\multicolumn{2}{p{0.76\textwidth}@{}}{\raggedright
\makecell[l]{
Speech Generation Expert: Hidden 2048, 4 layers, FFN 6144 (SwiGLU), RoPE \\
LLM-to-Expert Projector: 2-layer MLP, 4096 $\rightarrow$ 6144 $\rightarrow$ 2048, SiLU \\
RCP: Hidden 1024, 5 layers
}
} \\
\midrule 

Speech Codec &
\multicolumn{2}{p{0.76\textwidth}@{}}{\raggedright
\makecell[l]{
Mimi conv-transformer autoencoder; hidden 512; codebook 2048; 16/32 RVQ groups \\
12.5 Hz; causal SWA (10 s window)
}
} \\
\midrule 

Speech Output Adaptor &
\multicolumn{2}{p{0.76\textwidth}@{}}{\raggedright
2-layer MLP, 512 $\rightarrow$ 4096, GELU; post-RMSNorm (init=0.02)
} \\
\midrule 

Speaker Conditioning &
\multicolumn{2}{p{0.76\textwidth}@{}}{\raggedright
Pretrained frozen ECAPA-TDNN~\citep{dawalatabad2021ecapa}, followed by a linear projection, 192 $\rightarrow$ 4096
} \\

\midrule
\multicolumn{3}{c}{\textbf{Variant-specific modules}} \\
\midrule

Speech Encoder &
AuT Encoder~\citep{qwen3_asr_technical_report} &
Voxtral Realtime Encoder~\citep{voxtral_realtime} \\
\midrule 

Encoder Architecture &
\makecell[l]{
Hidden 1024, 24 layers, FFN 4096 (GELU) \\
Output 2048
} &
\makecell[l]{
Hidden 1280, 32 layers, FFN 5120 (SiLU) \\
Output 5120 (1280 $\times$ 4 frame stacking)
} \\
\midrule 

Attention Pattern &
Non-causal, full-context &
Causal SWA (15 s window) \\
\midrule 

Feature Extractor &
\makecell[l]{
Conv downsampling \\
Downsample hidden 480
} &
\makecell[l]{
Mel spectrogram \\
followed by patch embedding
} \\
\midrule 

Speech Input Adaptor &
\makecell[l]{
2-layer MLP, 2048 $\rightarrow$ 4096, GELU \\
post-RMSNorm (init=0.02)
} &
\makecell[l]{
2-layer MLP, 5120 $\rightarrow$ 4096, GELU \\
post-RMSNorm (init=0.02)
} \\

\bottomrule
\end{tabular}
\end{table*}
Table~\ref{tab:arch_compact} summarizes the detailed architectural configurations of \RaonSpeech and \RaonChat.

\section{Full-Duplex Synthetic Data Generation Pipeline}
\label{app:fd_syc_pipeline}

To synthesize diverse conversations with natural interaction patterns for full-duplex data generation, we design a four-stage pipeline.

\paragraph{Stage 1: Dialogue generation.}

We first define high-level dialogue scenarios and then synthesize multi-turn dialogues using a Qwen3-based LLM~\citep{yang2025qwen3}. These scenarios are organized into three template families: task-oriented settings with domain-specific personas, open-domain conversations, and speech-game interactions. For task-oriented data, to increase the diversity of synthesized dialogues, we design 15 scenarios and vary the system prompt across three levels of specificity: \textit{minimal}, \textit{topic-guided}, and \textit{detailed}. For realistic conversational dynamics, we further incorporate both direct flows, where the assistant responds immediately, and inquiry flows, where clarification is required. In addition, we include 7 speech-game settings, some of which naturally induce simultaneous speech.

\paragraph{Stage 2: Timeline construction and TTS synthesis.}
The generated dialogues are converted into dual-channel timelines with sample-accurate timestamps at 24\,kHz. Backchannel and interruption events are placed according to their annotated conversational roles, yielding overlapping speech patterns appropriate for full-duplex interaction. The resulting utterances are then synthesized with Qwen3-TTS~\citep{hu2026qwen3tts} using speaker-conditioned speech synthesis.

\paragraph{Stage 3: Timing refinement.}
To improve the naturalness of interaction timing, we refine the initial text-anchored backchannel positions using an audio-based backchannel prediction model. In contrast, simultaneous speech and barge-in events are handled with rule-based timing strategies. We further filter low-confidence backchannels and optionally insert new ones at detected backchannel opportunities.

\paragraph{Stage 4: Barge-in text truncation.}
For barge-in turns, we truncate the assistant audio at a randomly selected point within the utterance to simulate natural interruption. We then apply a forced-alignment model to locate the nearest word boundary and truncate the transcript accordingly, ensuring consistency between the text and the audible portion of the utterance.

\section{Detailed Spoken Question Answering Results}
\label{app:spokenqa_details}

For readability, the main English and Korean summary tables report aggregate VoiceBench/OpenAudioBench and KVoiceBench/KOpenAudioBench scores. The tables in this section unpack those four aggregate rows into their per-benchmark results. They do not change the total benchmark count in the paper, which remains 42 after excluding the aggregate summary rows.

\subsection{English Spoken Question Answering}
Table~\ref{tab:eng_spokenqa_details} unpacks the aggregate VoiceBench and OpenAudioBench rows from Table~\ref{tab:eng_speechlm_performance}. The VoiceBench block covers OpenBookQA, MMSU, BBH, SD-QA, AlpacaEval, CommonEval, WildVoice, IFEval, and AdvBench, whereas the OpenAudioBench block covers AlpacaEval, LlamaQ, TriviaQA, and WebQ. AlpacaEval appears in both blocks because we keep the benchmark groupings used for the main aggregate rows. The Average rows reproduce the aggregate scores reported in the main table.

\definecolor{lightgray}{gray}{0.90}
\providecommand{\best}[1]{\textbf{#1}}
\providecommand{\secondbest}[1]{\underline{#1}}

\begin{table*}[htbp]
\centering
\caption{Detailed English spoken question answering results for \RaonSpeech. The upper block unpacks the aggregate VoiceBench row and the lower block unpacks the aggregate OpenAudioBench row from the main table. \best{Bold} and \secondbest{underline} indicate the best and the second-best performance, respectively.}
\label{tab:eng_spokenqa_details}
\small
\renewcommand{\arraystretch}{1.08}
\resizebox{\textwidth}{!}{
\begin{tabular}{l>{\columncolor{blue!10}}c cccccccc}
\toprule
\textbf{Benchmark} & \textbf{Raon} & \textbf{Qwen2.5} & \textbf{Kimi} & \textbf{Audio} & \textbf{Step-Audio} & \textbf{Interactive} & \textbf{Fun-Audio} & \textbf{HyperCLOVA} & \textbf{MiniCPM} \\
 & \textbf{-Speech} & \textbf{-Omni} & \textbf{-Audio} & \textbf{Flamingo3} & \textbf{2 mini} & \textbf{Omni} & \textbf{Chat} & \textbf{X 8B Omni} & \textbf{-o 4.5}\\
\midrule
\rowcolor{lightgray}
\multicolumn{10}{c}{\textbf{VoiceBench} $\uparrow$} \\
\midrule
OpenBookQA & \secondbest{86.15} & 83.30 & 83.96 & 62.42 & 77.14 & 76.48 & 81.76 & 29.01 & \best{87.69} \\
MMSU & \best{67.50} & 56.31 & 59.82 & 48.60 & 54.49 & 58.72 & 65.29 & 30.12 & \secondbest{66.69} \\
BBH & \best{85.10} & 52.90 & 61.80 & 29.20 & 45.90 & 39.70 & \secondbest{67.70} & 37.90 & 55.00 \\
SD-QA & 60.94 & 54.07 & 59.49 & 38.16 & 36.35 & 43.58 & \secondbest{61.66} & 42.68 & \best{68.35} \\
AlpacaEval & 77.4 & 72.4 & 70.6 & 36.2 & 60.6 & 77.0 & \secondbest{79.4} & 63.2 & \best{84.0} \\
CommonEval & \secondbest{69.2} & 68.0 & 68.2 & 50.4 & 32.4 & 70.0 & 68.6 & 62.4 & \best{71.8} \\
WildVoice & \secondbest{70.0} & 62.6 & 60.4 & 42.6 & 41.8 & 63.2 & 66.6 & 51.8 & \best{71.6} \\
IFEval & \secondbest{78.06} & 51.40 & 56.18 & 18.13 & 47.53 & 45.13 & 76.56 & 35.39 & \best{80.59} \\
AdvBench & 96.73 & \secondbest{99.42} & \best{99.81} & 48.65 & 56.15 & 87.88 & 95.19 & 85.77 & 98.85 \\
\midrule
Average & \best{76.79} & 66.71 & 68.92 & 41.60 & 50.26 & 62.41 & 73.64 & 48.70 & \secondbest{76.06} \\
\midrule
\rowcolor{lightgray}
\multicolumn{10}{c}{\textbf{OpenAudioBench} $\uparrow$} \\
\midrule
AlpacaEval & 77.4 & 72.4 & 70.6 & 36.2 & 60.6 & 77.0 & \secondbest{79.4} & 63.2 & \best{84.0} \\
LlamaQ & \best{81.33} & 77.00 & \best{81.33} & 69.00 & 72.33 & 78.00 & \secondbest{80.67} & 76.67 & \secondbest{80.67} \\
TriviaQA & 62.00 & 58.50 & 58.10 & 28.50 & 53.20 & 57.10 & \secondbest{67.30} & 44.30 & \best{68.70} \\
WebQ & 60.10 & 59.00 & \secondbest{62.90} & 21.80 & 52.40 & 54.60 & 62.20 & 45.60 & \best{65.90} \\
\midrule
Average & 70.21 & 66.73 & 68.23 & 38.88 & 59.63 & 66.68 & \secondbest{72.39} & 57.44 & \best{74.82} \\
\bottomrule
\end{tabular}
}
\end{table*}

\subsection{Korean Spoken Question Answering}
Table~\ref{tab:kor_spokenqa_details} unpacks the aggregate KVoiceBench and KOpenAudioBench rows from Table~\ref{tab:kor_speechlm_performance}. The KVoiceBench block covers KOpenBookQA, KMMSU, KBBH, KSD-QA, KAlpacaEval, KCommonEval, KWildVoice, KIFEval, and KAdvBench, whereas the KOpenAudioBench block covers KAlpacaEval, KLlamaQ, KTriviaQA, and KWebQ. KAlpacaEval appears in both blocks because we keep the benchmark groupings used for the main aggregate rows. The Average rows reproduce the aggregate scores reported in the main table.

\definecolor{lightgray}{gray}{0.90}
\providecommand{\best}[1]{\textbf{#1}}
\providecommand{\secondbest}[1]{\underline{#1}}

\begin{table*}[htbp]
\centering
\caption{Detailed Korean spoken question answering results for \RaonSpeech. The upper block unpacks the aggregate KVoiceBench row and the lower block unpacks the aggregate KOpenAudioBench row from the main table. \best{Bold} and \secondbest{underline} indicate the best and the second-best performance, respectively.}
\label{tab:kor_spokenqa_details}
\small
\renewcommand{\arraystretch}{1.08}
\resizebox{\textwidth}{!}{
\begin{tabular}{l>{\columncolor{blue!10}}c ccccccc}
\toprule
\textbf{Benchmark} & \textbf{Raon} & \textbf{Qwen2.5} & \textbf{Audio} & \textbf{Step-Audio} & \textbf{Interactive} & \textbf{Fun-Audio} & \textbf{HyperCLOVA} & \textbf{MiniCPM} \\
 & \textbf{-Speech} & \textbf{-Omni} & \textbf{Flamingo3} & \textbf{2 mini} & \textbf{Omni} & \textbf{Chat} & \textbf{X 8B Omni} & \textbf{-o 4.5}\\
\midrule
\rowcolor{lightgray}
\multicolumn{9}{c}{\textbf{KVoiceBench} $\uparrow$} \\
\midrule
KOpenBookQA & \best{74.83} & \secondbest{31.01} & 6.52 & 7.42 & 7.42 & 28.31 & 27.42 & 25.17 \\
KMMSU & \best{55.46} & 28.34 & 10.01 & 11.68 & 7.02 & \secondbest{28.68} & 23.95 & 26.85 \\
KBBH & \best{83.47} & 49.87 & 36.27 & 40.27 & 8.27 & 52.27 & \secondbest{52.67} & 59.07 \\
KSD-QA & \best{44.84} & 27.95 & 7.50 & 23.26 & 5.63 & \secondbest{32.08} & 29.08 & 24.77 \\
KAlpacaEval & \best{65.0} & 56.0 & 29.4 & 43.4 & 26.8 & \secondbest{61.4} & 54.8 & 51.6 \\
KCommonEval & \secondbest{57.4} & \best{58.8} & 27.2 & 32.4 & 22.8 & 56.8 & 54.2 & 43.6 \\
KWildVoice & \best{59.6} & 50.6 & 26.0 & 35.2 & 23.0 & \secondbest{53.4} & 48.0 & 41.0 \\
KIFEval & \best{71.62} & 42.91 & 20.82 & 30.35 & 10.70 & \secondbest{53.35} & 32.62 & 34.59 \\
KAdvBench & \secondbest{87.33} & \best{95.91} & 5.65 & 64.33 & 68.03 & 84.80 & 83.24 & 48.54 \\
\midrule
Average & \best{66.62} & 49.04 & 18.82 & 32.03 & 19.96 & \secondbest{50.12} & 45.11 & 39.47 \\
\midrule
\rowcolor{lightgray}
\multicolumn{9}{c}{\textbf{KOpenAudioBench} $\uparrow$} \\
\midrule
KAlpacaEval & \best{65.0} & 56.0 & 29.4 & 43.4 & 26.8 & \secondbest{61.4} & 54.8 & 51.6 \\
KLlamaQ & \best{62.68} & 47.89 & 11.27 & 35.92 & 9.51 & 51.76 & \secondbest{60.21} & 38.38 \\
KTriviaQA & \best{35.47} & 17.99 & 4.45 & 17.37 & 3.62 & 23.68 & \secondbest{26.89} & 22.65 \\
KWebQ & \best{45.26} & 35.05 & 5.26 & 27.32 & 5.88 & 35.36 & \secondbest{38.45} & 30.00 \\
\midrule
Average & \best{52.10} & 39.23 & 12.60 & 31.00 & 11.45 & 43.05 & \secondbest{45.09} & 35.66 \\
\bottomrule
\end{tabular}
}
\end{table*}

\section{Speech Understanding Capability-Wise Results}
\label{app:audio_understanding_groups}

For the condensed Korean summary tables, KMMAU is reported using the average of the capability-wise accuracies. The detailed capability-level baseline results used for those summary rows are listed in Table~\ref{tab:audio_understanding_groups_ko}.

\providecommand{\best}[1]{\textbf{#1}}
\providecommand{\secondbest}[1]{\underline{#1}}
\begin{table*}[htbp]
\centering
\caption{Korean speech understanding capability accuracies for KMMAU. \best{Bold} and \secondbest{underline} indicate the best and the second-best performance, respectively.}
\label{tab:audio_understanding_groups_ko}
\small
\renewcommand{\arraystretch}{1.08}
\setlength{\tabcolsep}{5pt}
\resizebox{\textwidth}{!}{
\begin{tabular}{l c c c c c c c c}
\toprule
\textbf{Capability} & \cellcolor{blue!10}\textbf{\makecell{Raon-\\Speech}} & \textbf{\makecell{Qwen2.5-\\Omni}} & \textbf{\makecell{Audio\\Flamingo 3}} & \textbf{\makecell{Step-Audio\\2 mini}} & \textbf{\makecell{Interactive\\Omni}} & \textbf{\makecell{Fun-Audio-\\Chat}} & \textbf{\makecell{HyperCLOVA X\\8B Omni}} & \textbf{\makecell{MiniCPM-o\\4.5}} \\
\midrule
\# Speakers   & \cellcolor{blue!10} 30.00 & 29.00 & 26.00 & 28.00 & 23.00 & \secondbest{33.00} & 22.00 & \best{36.00} \\
Age            & \cellcolor{blue!10} \best{61.96} & 41.30 & 21.01 & 52.17 & 28.99 & 47.46 & 18.00 & \secondbest{51.81} \\
Gender         & \cellcolor{blue!10} \secondbest{93.70} & \best{98.89} & 57.04 & 88.89 & 46.30 & 87.78 & 30.00 & 91.11 \\
Fact Extraction & \cellcolor{blue!10} 86.87 & 81.82 & 59.60 & 79.80 & 33.33 & \best{91.92} & 34.34 & \secondbest{88.89} \\
General Counting & \cellcolor{blue!10} \secondbest{65.38} & 38.46 & 34.62 & 40.38 & 25.00 & \best{69.23} & 26.92 & 42.31 \\
Topic Summary  & \cellcolor{blue!10} 92.00 & 82.00 & 72.00 & 91.00 & 32.00 & \secondbest{96.00} & 70.00 & \best{98.00} \\
Role / Profession & \cellcolor{blue!10} 75.00 & 80.00 & 69.00 & \secondbest{83.00} & 14.00 & \best{92.00} & 44.00 & 78.00 \\
Word Frequency & \cellcolor{blue!10} \best{42.78} & 34.44 & 13.89 & 25.56 & 21.11 & \secondbest{42.22} & 18.89 & 26.11 \\
Word Order     & \cellcolor{blue!10} \best{98.75} & 80.37 & 54.09 & \secondbest{94.03} & 52.55 & 74.11 & 16.00 & 54.28 \\
\midrule
Average       & \cellcolor{blue!10} \best{71.83} & 62.92 & 45.52 & 64.76 & 30.70 & \secondbest{70.41} & 31.13 & 62.95 \\
\bottomrule
\end{tabular}
}
\end{table*}

\section{Full-Duplex Evaluation Details}
\label{app:fd_eval_details}

\providecommand{\fdpending}{\textcolor{BrickRed}{TBD}}
\providecommand{\fdpendingtext}[1]{\textcolor{BrickRed}{#1}}
\providecommand{\fdbest}[1]{\textbf{#1}}

We use the official FDB benchmark sets from Full-Duplex-Bench v1.0, v1.5, and v2.0~\citep{fdbv1,fdbv15,fdbv2}. For the offline FDB v1.0 and v1.5 evaluations, we use an internal pipeline based on the official benchmark definitions and reference implementation, and summarize the main implementation differences relevant to the reported scores below. The current FDB v2.0 session summary covers 72 multi-turn sessions per completed baseline.
\paragraph{Evaluator differences.}
The main implementation differences that affect the reported metrics are as follows.
\begin{itemize}
    \item Pause handling uses turn-based takeover detection with a 1.5-second and 5-word threshold, whereas the public v1.0 scripts use chunk-level rules with a 1.0-second and 3-word threshold.
    \item Smooth turn-taking and user interruption use automatic speech recognition (ASR) refined anchors, a 0.5-second post-anchor margin, and latency clipping that maps negative values to zero.
    \item The FDB v1.5 overlap-timing evaluation uses metadata together with ASR-refined anchors when metadata are available.
    \item Behavior and user-interruption judgments are produced with GPT-5.2 rather than the judge models used in the public reference implementation.
\end{itemize}

\paragraph{Scenario-wise behavior distributions.}
The main table reports only the scenario-wise target behavior and the Unknown rate for FDB v1.5. Table~\ref{tab:fd_behavior_appendix} provides the full four-way behavior distributions for the reproduced baselines. For User Interruption, the desired category is Respond. For User Backchannel, Background Speech, and Talking to Others, the desired category is Resume. Unknown is also reported because the official annotation protocol requires a valid new post-overlap segment after the overlap begins.

\begin{table*}[htbp]
\centering
\caption{Scenario-wise FDB v1.5 behavior distributions. \best{Bold} and \secondbest{underline} indicate the best and the second-best performance, respectively.}
\label{tab:fd_behavior_appendix}
\small
\renewcommand{\arraystretch}{1.08}
\setlength{\tabcolsep}{5pt}

\begin{tabular}{>{\raggedright\arraybackslash}p{2.2cm}
                >{\raggedright\arraybackslash}p{2.6cm}
                >{\centering\arraybackslash}p{1.35cm}
                >{\centering\arraybackslash}p{1.35cm}
                >{\centering\arraybackslash}p{1.45cm}
                >{\centering\arraybackslash}p{1.35cm}}
\toprule
\textbf{Scenario} & \textbf{Model} & \textbf{Respond} & \textbf{Resume} & \textbf{Uncertain} & \textbf{Unknown} \\
\midrule
\multirow{5}{*}{\makecell[l]{User\\Backchannel}} & Moshi & 0.010 & 0.092 & 0.000 & 0.898 \\
 & Freeze-Omni & 0.010 & \fdsecondbest{0.480} & 0.020 & \fdsecondbest{0.490} \\
 & PersonaPlex & 0.020 & 0.418 & 0.041 & 0.520 \\
 & MiniCPM-o 4.5 & 0.000 & \fdbest{0.520} & 0.000 & \fdbest{0.480} \\
 & \cellcolor{blue!10}\textbf{Raon-SpeechChat} & \cellcolor{blue!10}0.010 & \cellcolor{blue!10}0.398 & \cellcolor{blue!10}0.010 & \cellcolor{blue!10}0.582 \\
\midrule
\multirow{5}{*}{\makecell[l]{Background\\Speech}} & Moshi & 0.210 & 0.100 & 0.030 & 0.660 \\
 & Freeze-Omni & 0.770 & 0.100 & 0.000 & \fdbest{0.130} \\
 & PersonaPlex & 0.220 & 0.160 & 0.110 & 0.510 \\
 & MiniCPM-o 4.5 & 0.460 & \fdbest{0.260} & 0.000 & 0.280 \\
 & \cellcolor{blue!10}\textbf{Raon-SpeechChat} & \cellcolor{blue!10}0.530 & \cellcolor{blue!10}\fdsecondbest{0.230} & \cellcolor{blue!10}0.020 & \cellcolor{blue!10}\fdsecondbest{0.220} \\
\midrule
\multirow{5}{*}{\makecell[l]{Talking to\\Others}} & Moshi & 0.210 & \fdbest{0.210} & 0.030 & 0.550 \\
 & Freeze-Omni & 0.670 & \fdsecondbest{0.150} & 0.000 & \fdbest{0.180} \\
 & PersonaPlex & 0.310 & 0.120 & 0.200 & 0.370 \\
 & MiniCPM-o 4.5 & 0.550 & 0.130 & 0.010 & 0.310 \\
 & \cellcolor{blue!10}\textbf{Raon-SpeechChat} & \cellcolor{blue!10}0.620 & \cellcolor{blue!10}\fdsecondbest{0.150} & \cellcolor{blue!10}0.040 & \cellcolor{blue!10}\fdsecondbest{0.190} \\
\midrule
\multirow{5}{*}{\makecell[l]{User\\Interruption}} & Moshi & 0.560 & 0.145 & 0.020 & 0.275 \\
 & Freeze-Omni & \fdbest{0.810} & 0.085 & 0.015 & \fdsecondbest{0.090} \\
 & PersonaPlex & 0.710 & 0.100 & 0.075 & 0.115 \\
 & MiniCPM-o 4.5 & 0.660 & 0.220 & 0.005 & 0.115 \\
 & \cellcolor{blue!10}\textbf{Raon-SpeechChat} & \cellcolor{blue!10}\fdsecondbest{0.725} & \cellcolor{blue!10}0.140 & \cellcolor{blue!10}0.050 & \cellcolor{blue!10}\fdbest{0.085} \\
\bottomrule
\end{tabular}

\end{table*}

\paragraph{Scenario-wise latency with denominator.}
Table~\ref{tab:fd_latency_appendix} reports FDB v1.5 stop and response latency by scenario. Each latency is reported as a conditional mean over samples with a defined positive latency, and each denominator is shown explicitly as $n/N$. Stop latency is preferred to be higher for User Backchannel, Background Speech, and Talking to Others, but lower for User Interruption. Response latency is lower-is-better in all four scenarios. The Stop $n/N$ and Resp.\ $n/N$ columns are independent valid-measurement coverages rather than disjoint partitions, so their sum can exceed $N$, and some samples contribute to neither column when no valid stop or post-overlap response event is detected.

\begin{table*}[htbp]
\centering
\caption{Scenario-wise FDB v1.5 latency details. Bold and underline indicate the best and second-best values, respectively, for the preferred Stop and Response latencies in each scenario.}
\label{tab:fd_latency_appendix}
\small
\renewcommand{\arraystretch}{1.08}
\setlength{\tabcolsep}{5pt}

\begin{tabular}{>{\raggedright\arraybackslash}p{2.2cm}
                >{\raggedright\arraybackslash}p{2.6cm}
                >{\centering\arraybackslash}p{1.25cm}
                >{\centering\arraybackslash}p{1.15cm}
                >{\centering\arraybackslash}p{1.35cm}
                >{\centering\arraybackslash}p{1.15cm}}
\toprule
\textbf{Scenario} & \textbf{Model} & \textbf{Stop} & \textbf{\makecell[c]{Stop\\$n/N$}} & \textbf{Resp.} & \textbf{\makecell[c]{Resp.\\$n/N$}} \\
\midrule
\multirow{5}{*}{\makecell[l]{User\\Backchannel}} & Moshi & 2.797 & 12/98 & 1.895 & 11/98 \\
 & Freeze-Omni & 4.106 & 86/98 & 1.600 & 38/98 \\
 & PersonaPlex & 3.867 & 61/98 & \fdsecondbest{1.307} & 38/98 \\
 & MiniCPM-o 4.5 & \fdsecondbest{4.130} & 91/98 & \fdbest{0.154} & 7/98 \\
 & \cellcolor{blue!10}\textbf{Raon-SpeechChat} & \cellcolor{blue!10}\fdbest{4.648} & \cellcolor{blue!10}81/98 & \cellcolor{blue!10}1.897 & \cellcolor{blue!10}22/98 \\
\midrule
\multirow{5}{*}{\makecell[l]{Background\\Speech}} & Moshi & 2.199 & 15/100 & 0.858 & 17/100 \\
 & Freeze-Omni & \fdbest{5.563} & 79/100 & 0.706 & 27/100 \\
 & PersonaPlex & 2.078 & 33/100 & \fdsecondbest{0.584} & 56/100 \\
 & MiniCPM-o 4.5 & 2.188 & 72/100 & 0.630 & 48/100 \\
 & \cellcolor{blue!10}\textbf{Raon-SpeechChat} & \cellcolor{blue!10}\fdsecondbest{2.533} & \cellcolor{blue!10}59/100 & \cellcolor{blue!10}\fdbest{0.138} & \cellcolor{blue!10}83/100 \\
\midrule
\multirow{5}{*}{\makecell[l]{Talking to\\Others}} & Moshi & \fdsecondbest{3.664} & 21/100 & 1.209 & 10/100 \\
 & Freeze-Omni & \fdbest{5.673} & 94/100 & 1.004 & 24/100 \\
 & PersonaPlex & 2.127 & 40/100 & \fdsecondbest{0.214} & 61/100 \\
 & MiniCPM-o 4.5 & 2.200 & 86/100 & 0.654 & 55/100 \\
 & \cellcolor{blue!10}\textbf{Raon-SpeechChat} & \cellcolor{blue!10}1.403 & \cellcolor{blue!10}70/100 & \cellcolor{blue!10}\fdbest{0.138} & \cellcolor{blue!10}83/100 \\
\midrule
\multirow{5}{*}{\makecell[l]{User\\Interruption}} & Moshi & 2.845 & 93/200 & 0.770 & 64/200 \\
 & Freeze-Omni & 5.307 & 159/200 & 0.814 & 104/200 \\
 & PersonaPlex & \fdbest{1.194} & 120/200 & \fdsecondbest{0.484} & 142/200 \\
 & MiniCPM-o 4.5 & 2.483 & 176/200 & 0.564 & 132/200 \\
 & \cellcolor{blue!10}\textbf{Raon-SpeechChat} & \cellcolor{blue!10}\fdsecondbest{1.426} & \cellcolor{blue!10}117/200 & \cellcolor{blue!10}\fdbest{0.210} & \cellcolor{blue!10}156/200 \\
\bottomrule
\end{tabular}

\end{table*}

\end{document}